\documentclass[10pt]{article} 
\usepackage[preprint]{tmlr}

\usepackage{amsmath,amsfonts,bm}

\def\eqref#1{equation~\ref{#1}}

\def\1{\bm{1}}

\DeclareMathAlphabet{\mathsfit}{\encodingdefault}{\sfdefault}{m}{sl}
\SetMathAlphabet{\mathsfit}{bold}{\encodingdefault}{\sfdefault}{bx}{n}

\usepackage{hyperref}
\usepackage{url}
\usepackage[utf8]{inputenc}
\usepackage[T1]{fontenc}

\usepackage{xspace}
\usepackage{graphicx}
\usepackage{hyperref}
\usepackage{amsmath}
\usepackage{amsfonts}
\usepackage[table]{xcolor}
\usepackage{amssymb}
\usepackage{amsthm}
\hypersetup{hidelinks}

\usepackage{algorithm}

\usepackage{algorithmic}

\usepackage{booktabs}
\usepackage{multirow}
\usepackage{subcaption}
\usepackage{graphicx}
\usepackage{adjustbox}
\usepackage{fancyhdr}
\usepackage{tablefootnote}
\usepackage{pgfplots}
\usepgfplotslibrary{groupplots}
\usetikzlibrary{calc}
\pgfplotsset{compat=1.18}
\usepackage{nicefrac}
\usepackage{microtype}
\usepackage{titletoc}
\usepackage[most]{tcolorbox}
\fancypagestyle{firstpagestyle}{
    \fancyhf{}

    \fancyhead[L]{%
        \includegraphics[
            height=0.82cm,
            trim=70 60 70 60,
            clip
        ]{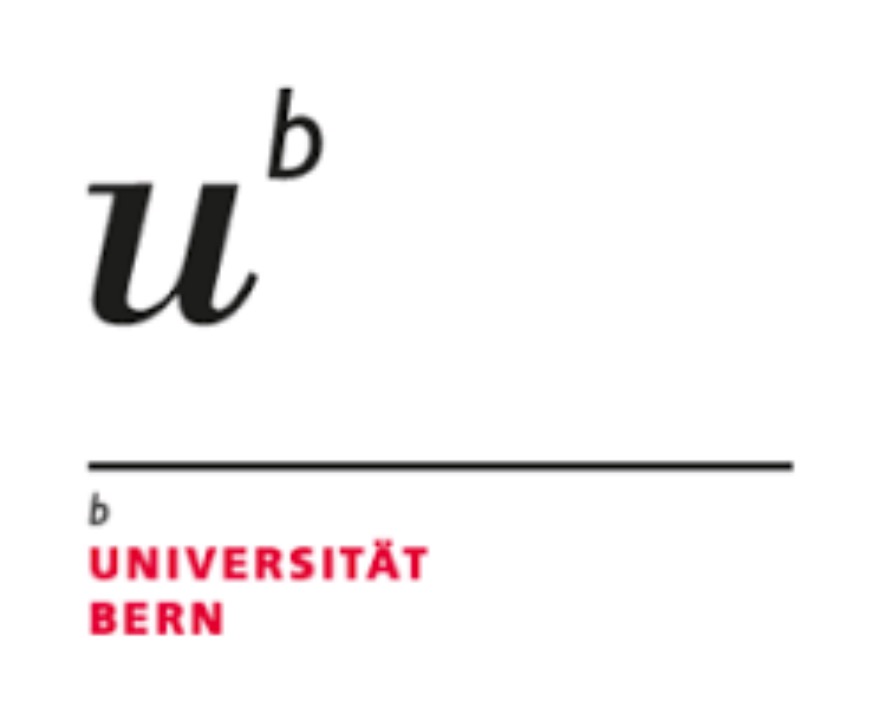}
    }

    \fancyhead[C]{%
        \includegraphics[
            width=3.35cm
        ]{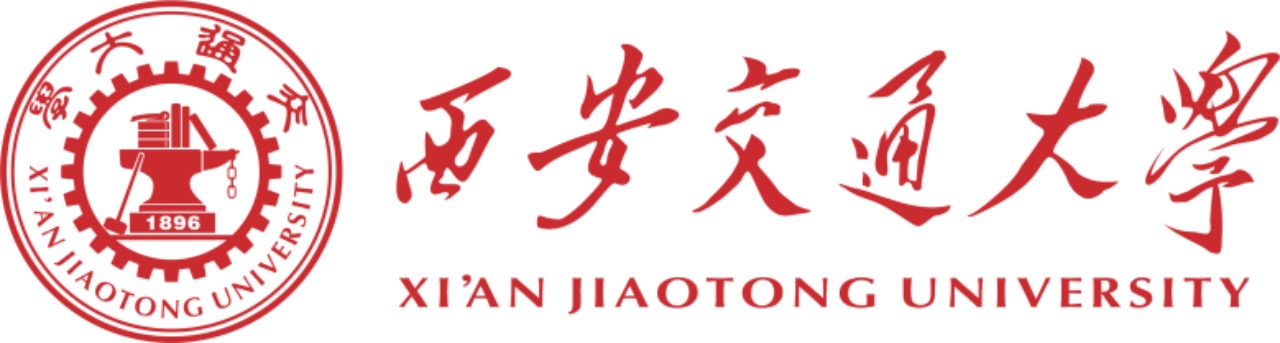}
    }

    \fancyhead[R]{%
        \includegraphics[
            height=0.68cm
        ]{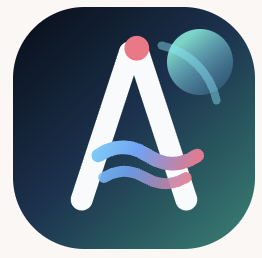}
    }

    \renewcommand{\headrulewidth}{0.4pt}
}

\usepackage{subcaption}

\usepackage[capitalize,noabbrev]{cleveref}

\theoremstyle{plain}
\newtheorem{theorem}{Theorem}[section]
\newtheorem{proposition}[theorem]{Proposition}

\theoremstyle{definition}

\theoremstyle{remark}

\definecolor{best}{RGB}{210, 235, 255}
\definecolor{second}{RGB}{255, 240, 210}
\definecolor{hotPink}{RGB}{200, 40, 120}
\definecolor{ObjBlue}{RGB}{36,82,139}
\definecolor{ObjOrange}{RGB}{196,104,0}

\newcommand{\best}[1]{\cellcolor{best}\textbf{#1}}
\newcommand{\second}[1]{\cellcolor{second}\textbf{#1}}

\newcommand{\regName}{\textsc{DAGR}$^{\dagger}$\xspace}

\newcommand{\chg}[1]{%
  \ifdim #1 pt > 0 pt
    {\textcolor{green!45!black}{+#1}}%
  \else
    {\textcolor{red!65!black}{#1}}%
  \fi
}

\newtcolorbox{propbox}{
    enhanced,
    colback=blue!2,
    colframe=blue!35!black,
    boxrule=0.6pt,
    arc=2pt,
    left=6pt,
    right=6pt,
    top=5pt,
    bottom=5pt,
    before skip=8pt,
    after skip=8pt
}

\newtcolorbox{contributionbox}{
    enhanced,
    colback=green!3,
    frame hidden,
    borderline west={2pt}{0pt}{green!45!black},
    sharp corners,
    left=8pt,
    right=7pt,
    top=7pt,
    bottom=7pt,
    before skip=8pt,
    after skip=8pt
}

\newtcolorbox{theorybox}{
    enhanced,
    breakable,
    colback=red!2,
    frame hidden,
    borderline west={2pt}{0pt}{red!45!black},
    sharp corners,
    left=8pt,
    right=7pt,
    top=7pt,
    bottom=7pt,
    before skip=9pt,
    after skip=9pt
}

\newtcolorbox{assumpbox}{
    enhanced,
    breakable,
    colback=yellow!7,
    colframe=yellow!45!black,
    boxrule=0.7pt,
    arc=2pt,
    left=6pt,
    right=6pt,
    top=5pt,
    bottom=5pt
}

\title{Diversity through Bounded Agreement: A Controlled Geometric Regularizer in Multimodal Learning}

\author{
\name
Zixuan Xia$^{1,3,*}$
\quad
Hao Wang$^{1,3,*}$
\quad
Pengcheng Weng$^{1,3,*}$
\quad
Yanyu Qian$^{2,3}$
\\[0.3em]
Yangxin Xu$^{3}$
\quad
William Dan$^{3}$
\quad
Fei Wang$^{3,\dagger}$
\\[0.7em]
\addr
$^{1}$Department of Informatics, University of Bern,
Bern, Switzerland
\\
$^{2}$College of Computing and Data Science,
Nanyang Technological University, Singapore
\\
$^{3}$AIRS Lab, School of Software Engineering,
Xi'an Jiaotong University, Xi'an, China
\\[0.4em]
{\rm
$^{*}$Equal contribution.
\qquad
$^{\dagger}$Corresponding author:
\href{mailto:feynmanw@xjtu.edu.cn}{feynmanw@xjtu.edu.cn}
}
}

\def\month{MM}  
\def\year{YYYY} 
\def\openreview{\url{https://openreview.net/forum?id=XXXX}} 

\begin{document}
\thispagestyle{firstpagestyle}

\maketitle

\vspace{-0.3em}
\begingroup
\renewcommand{\thefootnote}{}
\footnotetext{
The authors used ChatGPT as a general-purpose assistive tool for
language editing and manuscript preparation. The authors take full
responsibility for all scientific content, experiments, derivations,
and conclusions.
}
\endgroup

\begin{abstract}
Multimodal fusion is often treated as an optimization-balancing problem, yet balancing losses or gradients does not directly control representation geometry. We introduce \regName, a lightweight training-time regularizer that enforces bounded cross-modal agreement: paired normalized representations are penalized only when their distance exceeds a tolerance $\tau$, while a modality-wise uniformity term prevents representation concentration. Without architectural changes or inference-time overhead, \regName consistently improves multimodal performance across audio-visual, image-text, and RF-based benchmarks. Controlled studies further show that moderate tolerance outperforms both exact alignment and overly weak coupling, highlighting the benefit of explicitly regulating representation geometry.
\end{abstract}

\section{Introduction}
\label{sec:introduction}

Multimodal learning combines heterogeneous signals such as audio, vision, and
language, but different modalities often contribute unevenly during training,
leading dominant modalities to suppress weaker ones~\cite{baltruvsaitis2018multimodal,
wei2024mmpareto,chaudhuri2025closer}.
Existing methods mainly address this as an optimization-balancing problem
through objective reweighting or gradient modification~\cite{wei2025boosting,yu2020gradient}.
We instead ask a complementary question:
\emph{how much should multimodal representations agree?}

Balanced optimization does not uniquely determine representation geometry.
Within each modality, representations may become overly concentrated, while
paired modalities may either drift too far apart or be forced into excessive
alignment that suppresses modality-specific information.
We therefore introduce \textbf{bounded agreement}, which encourages paired
representations to remain sufficiently close without requiring maximal
alignment.

\begin{figure*}[t]
    \centering
    \includegraphics[width=0.92\textwidth]{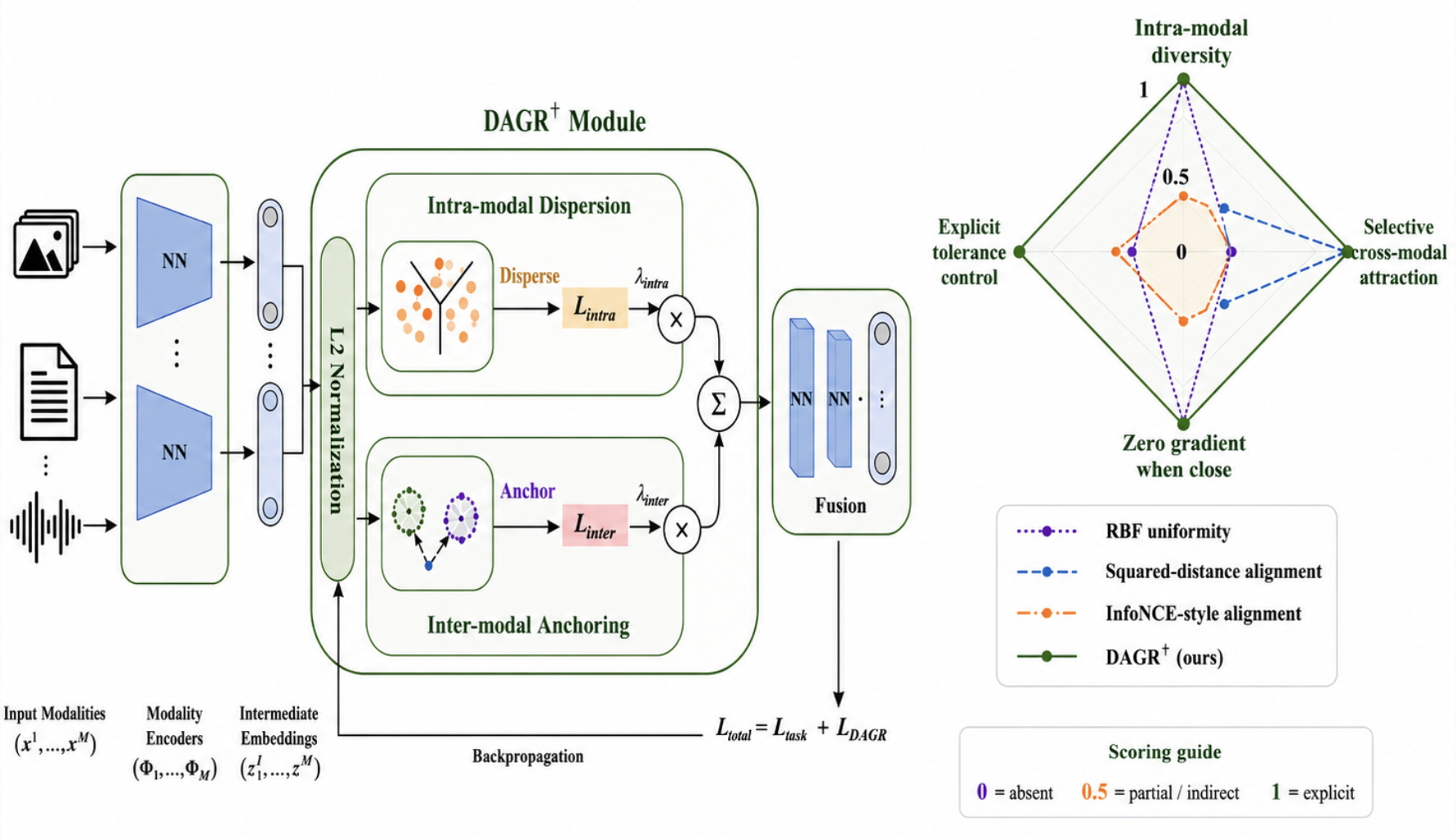}
    \caption{\textbf{Overview of the DAGR$^\dagger$ framework.}
    DAGR regularizes multimodal intermediate embeddings through
    intra-modal dispersion and inter-modal anchoring, and is optimized
    jointly with the task objective.}
    \label{fig:dagr_framework}
\end{figure*}

Based on this principle, we propose \textbf{\regName}
(\textbf{D}iversity--\textbf{A}greement \textbf{G}eometric
\textbf{R}egularization), a lightweight training-time regularizer operating on
normalized intermediate representations.
As illustrated in Fig.~\ref{fig:dagr_framework}, \regName combines a diversity
term that discourages intra-modal concentration with the bounded agreement loss
\begin{equation}
    \mathcal{L}_{\mathrm{a}}
    =
    \left[
    d(\mathbf{z}_i^m,\mathbf{z}_i^n)-\tau
    \right]_{+}^{2}.
\end{equation}
The agreement term is inactive when
$d(\mathbf{z}_i^m,\mathbf{z}_i^n)\leq\tau$, allowing modality-specific
structure to be preserved within the tolerance region.
\regName requires no architectural modification or additional inference-time
computation.

The tolerance $\tau$ controls the agreement regime: small values approach rigid alignment, intermediate values yield bounded agreement, and large values effectively remove the constraint. We evaluate whether this intermediate regime provides a better trade-off using controlled alignment comparisons and representation-geometry diagnostics
across diverse multimodal benchmarks.

\noindent\textbf{Our main contributions are as follows:}
\begin{itemize}
    \item We distinguish task-level or optimization-level objectives from explicit control of modality-wise representation geometry.
    \item We introduce \regName, a lightweight geometric regularization framework that preserves representation diversity while enforcing a zero-gradient agreement region, without architectural modification or inference-time overhead.
    \item We evaluate \regName across multimodal classification, clustering, and retrieval, together with controlled studies of modality imbalance, representation geometry, and correspondence corruption.
\end{itemize}

\section{Related Work}
\label{sec:related-work}

\subsection{Optimization and modality imbalance.}
Prior work on supervised multimodal learning largely addresses modality imbalance by modifying modality-specific optimization dynamics. Representative approaches rebalance logits or gradients~\cite{peng2022balanced,zong2024balancing,kwon2025see,wei2025boosting}, mitigate gradient conflicts through multi-objective optimization~\cite{sener2018multi,yu2020gradient,fernando2025mitigating,jiang2025interactive}, or use distillation, asymmetric optimization, and decoupled supervision to
prevent dominant modalities from suppressing weaker ones
~\cite{rakib2025g2d,wei2025imbalanced,yang2024facilitating,
zhang2024multimodal,wei2024diagnosing}.

These methods primarily regulate how learning signals are weighted or
coordinated across modalities.
Our work explores a complementary direction: instead of modifying optimization
dynamics, \regName directly regularizes intermediate representation geometry
through bounded cross-modal agreement.

\subsection{Multimodal alignment and shared representations.}
Another broad line of research learns shared representation spaces by promoting
semantic correspondence across modalities.
Contrastive pretraining and retrieval methods, including CLIP, ALIGN, FLAVA,
BLIP, ImageBind, and LanguageBind, attract paired observations while
distinguishing non-matching samples~\cite{radford2021learning,li2021align,
singh2022flava,li2022blip,girdhar2023imagebind,zhu2023languagebind}.
Related approaches explicitly separate shared and modality-specific factors
through invariant-specific decomposition or latent-variable
modeling~\cite{hazarika2020misa,wu2018multimodal,shi2019variational,
sutter2021generalized,palumbo2024deep,gaodisentangled}.
Collectively, these studies highlight two coupled objectives: preserving
cross-modal correspondence and retaining information that is specific to each
modality.
However, the appropriate strength and support of cross-modal alignment may
depend on the task, modality combination, and training setting.

\subsection{Representation geometry, uniformity, and bounded agreement.}
Alignment and uniformity provide a geometric perspective on representation
learning~\cite{wang2020understanding}, while variance-preserving and
redundancy-reducing objectives help avoid degenerate representations
~\cite{zbontar2021barlow,bardes2021vicreg}.
Recent multimodal studies further analyze modality gaps and the structures that
should be shared across modalities~\cite{dufumier2024align,yaras2024explaining,
yi2025decipher,chaudhuri2025closer}.

Most related to our formulation, UniAlign~\cite{yin2026towards} decouples
modality-wise uniformity from cross-modal alignment in multimodal InfoNCE.
Our focus is different: in supervised multimodal fusion, we ask whether
pairwise attraction should remain active once paired representations are
sufficiently close.
\regName combines modality-wise RBF uniformity with a bounded agreement term,
where $\tau=0$ recovers always-on alignment and $\tau>0$ introduces a tolerance
band.
Our experiments therefore isolate the effect of this bounded-agreement design
relative to exact alignment and related geometric controls.

\section{Methodology}
\label{sec:method}

\subsection{Problem Setup and Geometric Motivation}
\label{sec}

We consider multimodal learning over paired observations. Let
\begin{equation}
\mathcal{D}={x_i}_{i=1}^{N},
\qquad
x_i=(x_i^{1},\ldots,x_i^{M}),
\end{equation}
where $x_i^m$ denotes the observation from the modality $m$ associated
with the sample $i$. Depending on the application, the paired observations may also be accompanied by class labels, relevance annotations, or other task-specific supervision. At the layer where \regName is applied,each modality-specific encoder $\phi_m$ produces an intermediate
representation
\begin{equation}
z_i^m=\phi_m(x_i^m)\in\mathbb{R}^{d}.
\end{equation}
We assume that the selected intermediate representations share a common
dimensionality, so that their modality-wise and cross-modal geometric
relations can be evaluated in the same space. For example, in supervised multimodal fusion:
\begin{equation}
\mathbb{E}_{(x,y)\sim\mathcal{D}}
\left[
\ell\left(g(z^{1},\ldots,z^{M}),y\right)
\right].
\label{eq}
\end{equation}

Although the task objective constrains representations in ways that are
relevant to the target application, it needs no explicitly determine all
aspects of their geometry. For instance, under a linear prediction head,
transformations along directions invisible to the head can preserve the
logits while changing pairwise distances and covariance structure.
Proposition~\ref{prop:app_geom_ambiguity} in the Appendix provides a restricted illustration of this non-identifiability. This example is not intended as a general characterization of every task objective or multimodal architecture considered in this work.

More generally, task-specific objectives may shape representation
geometry only indirectly, or according to criteria different from the
geometric behavior considered here. In particular, they do not
necessarily combine explicit control of within-modality concentration
with a finite tolerance for paired cross-modal discrepancy. We therefore
augment the original task objective with two complementary geometric
preferences: representations within each modality should avoid excessive
concentration, while paired modalities should be prevented from drifting
excessively far apart without being forced to coincide. These preferences
lead to the modality-wise dispersion and bounded cross-modal agreement
terms introduced next.

\subsection{\regName: Diversity under Bounded Agreement}
\label{sec:dagr}

We instantiate this principle with \regName
(\underline{D}iversity-and-\underline{A}greement
\underline{G}eometric \underline{R}egularization),
a lightweight training-time regularizer consisting of two complementary
terms: modality-wise diversity and bounded cross-modal agreement.

\paragraph{Normalized representation space.}
Before computing the geometric regularizers, we normalize each
representation to the unit hypersphere,
$\tilde z_i^m=z_i^m/\|z_i^m\|_2$.
Normalization removes feature-scale variation from the regularized
geometry and is used only for computing \regName.

\paragraph{Diversity and bounded agreement.}
For a mini-batch $\mathcal{B}$ of size $B$, we define
\begin{align}
\mathcal{L}_{\mathrm{d}}
&=
\frac{1}{M}
\sum_{m=1}^{M}
\log
\left[
\frac{1}{B(B-1)}
\sum_{\substack{i,j\in\mathcal{B}\\i\neq j}}
\exp\!\left(
-t\|\tilde z_i^m-\tilde z_j^m\|_2^2
\right)
\right],
\label{eq:dagr_disp}
\\
\mathcal{L}_{\mathrm{a}}
&=
\frac{1}{B\binom{M}{2}}
\sum_{i\in\mathcal{B}}
\sum_{m<n}
\left[
\|\tilde z_i^m-\tilde z_i^n\|_2-\tau
\right]_+^2 ,
\label{eq:dagr_anchor_general}
\end{align}
where $t>0$ controls the interaction scale,
$(x)_+=\max(x,0)$, and $\tau\ge0$ is the admissible
cross-modal discrepancy.

$\mathcal{L}_{\mathrm{d}}$ is a standard RBF uniformity objective
applied separately to each modality.
Because the RBF interaction decays exponentially with pairwise distance,
minimizing $\mathcal{L}_{\mathrm{d}}$ primarily discourages locally
concentrated representations.
In contrast, $\mathcal{L}_{\mathrm{a}}$ operates on paired
cross-modal representations and penalizes only the portion of their
distance that exceeds $\tau$.

\paragraph{Training objective.}
The complete objective is
\begin{equation}
\mathcal{L}
=
\mathcal{L}_{\mathrm{task}}
+
\lambda_{\mathrm{d}}\mathcal{L}_{\mathrm{d}}
+
\lambda_{\mathrm{a}}\mathcal{L}_{\mathrm{a}},
\label{eq:full_objective}
\end{equation}
where $\lambda_{\mathrm{d}}$ and $\lambda_{\mathrm{a}}$ control the
strengths of the two geometric terms.
Because \regName acts only on intermediate representations during
training, it requires no architectural modification and introduces no
additional inference-time computation.

When adaptive weighting is enabled, the two auxiliary gradients are
balanced using a minimum-norm Pareto weighting scheme.
This is an optimization mechanism rather than a defining component of
\regName; its derivation and implementation details are provided in the
Appendix.

\subsection{Geometric Interpretation of Bounded Agreement}
\label{sec:properties}

The key distinction between bounded agreement and conventional
cross-modal alignment is that paired representations are not
continuously attracted toward coincidence.
Let
$d=\|\tilde z^m-\tilde z^n\|_2$
and consider the pairwise penalty
$\ell_{\mathrm{a}}(d)=[d-\tau]_+^2$.
Its derivative with respect to the paired distance is
\begin{equation}
\frac{\partial \ell_{\mathrm{a}}}{\partial d}
=
\begin{cases}
0, & d\le\tau,\\
2(d-\tau), & d>\tau.
\end{cases}
\label{eq:bounded_gradient}
\end{equation}

Hence, once a paired representation enters the tolerance band
$d\le\tau$, it receives no further agreement gradient.
This differs from rigid squared-distance alignment, which continues to
contract every non-identical pair.
The tolerance $\tau$ therefore specifies an admissible region rather
than a target distance.

Since all representations are normalized,
$0\le d\le2$.
Consequently, $\tau=0$ recovers always-active squared-distance
alignment, $0<\tau<2$ yields bounded agreement, and $\tau\ge2$
removes explicit cross-modal agreement regularization.
Together, $\mathcal{L}_{\mathrm{d}}$ and $\mathcal{L}_{\mathrm{a}}$
therefore regulate complementary aspects of representation geometry:
the former discourages excessive within-modality concentration, while
the latter limits excessive cross-modal drift without suppressing
modality-specific variation once sufficient agreement has been reached.

\section{Experiments and Results}
\subsection{Datasets}
We evaluate \regName on seven benchmarks spanning audio--visual recognition(CREMA-D and Kinetics-Sounds), language--audio--visual learning (URFunny),
image--text clustering (CUBICC), image--text retrieval (Flickr30K and COCO), and RF-based activity recognition (XRF55). We follow the standard evaluation protocols of the corresponding baseline methods; detailed dataset statistics, splits, and implementation settings are
provided in Appendix~\ref{app:implementation}.

\subsection{Main Results Across Multimodal Benchmarks}
\label{sec:main_results}

\begin{table*}[t]
\centering
\caption{
Main results on CREMA-D, Kinetics-Sounds, and URFunny.
We report unimodal and multimodal classification accuracy (\%) on CREMA-D and
Kinetics-Sounds (3-seeds, full mean and std see Table~\ref{tab:ablation_component}), and multimodal F1 (\%) on the tri-modal URFunny benchmark.
}
\label{tab:main_all}
\vspace{0.35em}

\footnotesize
\setlength{\tabcolsep}{3.6pt}
\renewcommand{\arraystretch}{1.08}

\begin{tabular*}{\textwidth}{@{\extracolsep{\fill}}lccccccc@{}}
\toprule
\multirow{2}{*}{Method}
& \multicolumn{3}{c}{CREMA-D}
& \multicolumn{3}{c}{Kinetics-Sounds}
& \multicolumn{1}{c}{URFunny} \\
\cmidrule(lr){2-4}
\cmidrule(lr){5-7}
\cmidrule(lr){8-8}
& Audio & Visual & Multi
& Audio & Visual & Multi
& Multi F1 \\
\midrule

G-Blending~\cite{wang2020makes}
& 58.78 & 58.62 & 69.21
& 46.35 & 51.12 & 69.60
& 63.08 \\

OGM-GE~\cite{peng2022balanced}
& 57.76 & 40.09 & 68.82
& 44.23 & 45.81 & 66.89
& 62.78 \\

PMR~\cite{fan2023pmr}
& 55.11 & 38.34 & 67.44
& 43.61 & 46.67 & 65.70
& 63.58 \\

AGM~\cite{li2023boosting}
& 56.37 & 43.54 & 69.61
& 46.12 & 47.65 & 68.88
& 63.68 \\

MMPareto~\cite{wei2024mmpareto}
& 59.43 & 61.09 & 70.12
& 48.40 & 52.42 & 69.83
& 64.99 \\

D\&R~\cite{wei2024diagnosing}
& 61.11 & 64.57 & 74.32
& 49.78 & 54.88 & 69.10
& 63.48 \\

DGL~\cite{wei2025boosting}
& \second{62.17}
& \second{70.31}
& \second{77.65}
& \second{50.63}
& \second{59.83}
& \second{75.44}
& \second{68.27} \\

\midrule

\textbf{DGL $+$ \regName}
& \best{62.98}
& \best{72.10}
& \best{79.40}
& \best{52.79}
& \best{63.25}
& \best{77.15}
& \best{69.11} \\

\textbf{Imp $\uparrow$}
& \textcolor{green!45!black}{+0.81}
& \textcolor{green!45!black}{+1.79}
& \textcolor{green!45!black}{+1.75}
& \textcolor{green!45!black}{+2.16}
& \textcolor{green!45!black}{+3.42}
& \textcolor{green!45!black}{+1.71}
& \textcolor{green!45!black}{+0.84} \\

\bottomrule
\end{tabular*}
\end{table*}

\subsubsection{Audio--visual and tri-modal recognition.}
Table~\ref{tab:main_all} summarizes the results on CREMA-D, Kinetics-Sounds, and URFunny. On the two audio--visual benchmarks, adding \regName to DGL increases multimodal accuracy by $1.75\%$ on CREMA-D and $1.71\%$ on Kinetics-Sounds. The corresponding unimodal branches also improve: audio and visual accuracy increase by $0.81\%$ and $1.79\%$ , respectively, on CREMA-D, and by $2.16\%$ and $3.42\%$, respectively, on Kinetics-Sounds. On the three-modality URFunny benchmark, adding \regName increases the F1 score of DGL from $68.27$ to $69.11$. These benchmark-level results indicate that \regName can be integrated into both two- and three-modality training pipelines. 

\subsubsection{Image--text benchmarks: clustering and CLIP-based retrieval}
We next evaluate whether \regName generalizes to image--text representation learning. We consider two complementary settings: clustering on CUBICC and image--text retrieval on Flickr30K and COCO.

\paragraph{Clustering on CUBICC} As shown in Table~\ref{tab:cubicc}, we report ACC, NMI, and ARI for image-only, caption-only, and joint representations. 
Compared to the original DCMEM baseline, DCMEM $+$ \regName consistently improves clustering performance across all settings. For joint representations, \regName improves ACC by $+3.5$, NMI by $+1.3$, and ARI by $+2.4$. Consistent gains are also observed for unimodal representations. These results suggest that \regName improves semantic organization by strengthening inter-class separation while preserving compact class structure.

\begin{table*}[t]
\centering
\caption{
Clustering results on the CUB Image-Captions for Clustering (CUBICC) dataset.
We report ACC, NMI, and ARI (\%) for image-only, caption-only, and joint representations.
}
\label{tab:cubicc}
\vspace{0.35em}

\footnotesize
\setlength{\tabcolsep}{4.8pt}
\renewcommand{\arraystretch}{1.08}

\begin{tabular*}{\textwidth}{@{\extracolsep{\fill}}lccccccccc@{}}
\toprule
\multirow{2}{*}{Method}
& \multicolumn{3}{c}{Image}
& \multicolumn{3}{c}{Caption}
& \multicolumn{3}{c}{Joint} \\
\cmidrule(lr){2-4}\cmidrule(lr){5-7}\cmidrule(lr){8-10}
& ACC & NMI & ARI
& ACC & NMI & ARI
& ACC & NMI & ARI \\
\midrule
MVAE~\cite{wu2018multimodal}
& 26.2 & 12.4 & 7.5
& 18.1 & 2.4  & 0.9
& 38.7 & 26.8 & 18.0 \\
MMVAE~\cite{shi2019variational}
& 23.1 & 12.1 & 6.1
& 14.5 & 1.3  & 0.1
& 15.8 & 1.5  & 0.2  \\
MoPoE~\cite{sutter2021generalized}
& 33.4 & 17.6 & 11.5
& 43.5 & 27.1 & 19.9
& 40.8 & 30.4 & 20.2 \\
MEME~\cite{joy2021learning}
& 44.8 & 43.4 & 28.4
& 36.3 & 29.5 & 18.6
& 19.8 & 4.8  & 2.1  \\
MMVAE+~\cite{palumbo2023mmvae+}
& 27.7 & 11.9 & 7.1
& 48.7 & 36.4 & 26.8
& 64.4 & 52.6 & 44.1 \\
CMVAE~\cite{palumbo2024deep}
& 67.7 & 58.3 & 47.4
& 65.1 & 53.3 & 42.7
& 73.7 & 67.4 & 57.2 \\
MMVM~\cite{sutter2024unity}
& 58.9 & 56.9 & 44.5
& 23.9 & 9.4  & 5.4
& 66.8 & 67.0 & 55.5 \\
MVP~\cite{gao2025deep}
& 64.1 & 53.8 & 41.8
& 48.5 & 34.4 & 26.1
& 61.1 & 55.6 & 44.0 \\
DCMEM~\cite{gaodisentangled}
& \second{87.5} & \second{79.3} & \second{73.6}
& \second{72.2} & \second{56.0} & \second{48.1}
& \second{86.7} & \second{78.4} & \second{72.2} \\
\midrule
\textbf{DCMEM $+$ \regName}
& \best{89.3} & \best{79.8} & \best{76.8}
& \best{74.0} & \best{56.5} & \best{48.6}
& \best{90.2} & \best{79.7} & \best{74.6} \\
\textbf{Imp $\uparrow$}
& \textcolor{green!45!black}{+1.8}
& \textcolor{green!45!black}{+0.5}
& \textcolor{green!45!black}{+3.2}
& \textcolor{green!45!black}{+1.8}
& \textcolor{green!45!black}{+0.5}
& \textcolor{green!45!black}{+0.5}
& \textcolor{green!45!black}{+3.5}
& \textcolor{green!45!black}{+1.3}
& \textcolor{green!45!black}{+2.4} \\
\bottomrule
\end{tabular*}
\end{table*}

\paragraph{CLIP-based image--text retrieval.}
We further evaluate \regName on Flickr30K and COCO using CLIP-based image--text retrieval models. As shown in Table~\ref{tab:clip_clean_retrieval}, adding \regName can improve all reported recall metrics for both retrieval
directions on Flickr30K and COCO. The gains are particularly pronounced at R@1, increasing image-to-text retrieval from 86.2 to 87.2 on Flickr30K and from 79.9 to 81.9 on COCO, and text-to-image retrieval from 72.9 to 75.1 and from 65.0 to 68.0,
respectively, showing that \regName can also be integrated into CLIP-style cross-modal representation learning beyond classification and clustering.
\begin{table*}[t]
\centering
\caption{
Image--text retrieval results ($\%$) on Flickr30K and COCO.
We report $R@\{1,5,10\}$ for image-to-text (I2T) and text-to-image (T2I) retrieval.
}
\label{tab:clip_clean_retrieval}
\vspace{0.25em}

\footnotesize
\setlength{\tabcolsep}{4.0pt}
\renewcommand{\arraystretch}{1.05}

\begin{tabular*}{\textwidth}{
@{\extracolsep{\fill}}
llcccccc@{}
}
\toprule
\multirow{2}{*}{Direction}
& \multirow{2}{*}{Method}
& \multicolumn{3}{c}{Flickr30K}
& \multicolumn{3}{c}{COCO} \\
\cmidrule(lr){3-5}
\cmidrule(lr){6-8}
& & R@1 & R@5 & R@10
& R@1 & R@5 & R@10 \\
\midrule

\multirow{3}{*}{I2T}
& CLIP
& 86.2 & 97.6 & 99.2
& 79.9 & 95.1 & 98.1 \\

& \textbf{CLIP $+$ \regName}
& \textbf{87.2} & \textbf{98.1} & \textbf{99.5}
& \textbf{81.9} & \textbf{96.2} & \textbf{98.8} \\

& \textbf{Imp $\uparrow$}
& \textcolor{green!45!black}{+1.0}
& \textcolor{green!45!black}{+0.5}
& \textcolor{green!45!black}{+0.3}
& \textcolor{green!45!black}{+2.0}
& \textcolor{green!45!black}{+1.1}
& \textcolor{green!45!black}{+0.7} \\

\midrule

\multirow{3}{*}{T2I}
& CLIP
& 72.9 & 92.3 & 96.0
& 65.0 & 90.3 & 98.1 \\

& \textbf{CLIP $+$ \regName}
& \textbf{75.1} & \textbf{93.6} & \textbf{96.9}
& \textbf{68.0} & \textbf{92.1} & \textbf{98.8} \\

& \textbf{Imp $\uparrow$}
& \textcolor{green!45!black}{+2.2}
& \textcolor{green!45!black}{+1.3}
& \textcolor{green!45!black}{+0.9}
& \textcolor{green!45!black}{+3.0}
& \textcolor{green!45!black}{+1.8}
& \textcolor{green!45!black}{+0.7} \\

\bottomrule
\end{tabular*}
\vspace{-1mm}
\end{table*}

\subsubsection{RF-based multimodal benchmark: XRF55}
We further evaluate \regName on XRF55 using two multimodal sensing frameworks: X-Fi~\cite{chen2024x} and COMPASS~\cite{wang2026compass}.
As shown in Table~\ref{tab:main_xrf55}, adding \regName improves recognition
accuracy across all evaluated unimodal, bimodal, and tri-modal configurations
under both frameworks.
The average accuracy increases from $74.26\%$ to $75.57\%$ for X-Fi and from
$79.55\%$ to $83.44\%$ for COMPASS, corresponding to gains of $+1.31$ and
$+3.89$ points, respectively.
These results further demonstrate that \regName can be integrated into
heterogeneous RF-based multimodal pipelines beyond conventional
audio--visual and image--text learning.

\begin{table*}[t]
\centering
\caption{
Accuracy (\%) on XRF55 under two multimodal learning frameworks.
We report the original \textbf{X-Fi} and \textbf{COMPASS} baselines and their
corresponding variants augmented with \regName.
\textbf{Imp$\uparrow$} denotes the absolute improvement after applying \regName,
and \textbf{Avg.} averages over all seven modality configurations.
}
\label{tab:main_xrf55}
\vspace{0.25em}

\footnotesize
\setlength{\tabcolsep}{4.2pt}
\renewcommand{\arraystretch}{1.05}

\begin{tabular*}{\textwidth}{@{\extracolsep{\fill}}lcccccc@{}}
\toprule
\multirow{2}{*}{Modality}
& \multicolumn{3}{c}{X-Fi}
& \multicolumn{3}{c}{COMPASS} \\
\cmidrule(lr){2-4}
\cmidrule(lr){5-7}
& Base & $+$\regName & $\mathrm{Imp}\uparrow$
& Base & $+$\regName & Imp$\uparrow$ \\
\midrule

mmWave
& 82.33
& \textbf{83.57}
& \textcolor{green!45!black}{+1.24}
& 88.55
& \textbf{90.56}
& \textcolor{green!45!black}{+2.01} \\

WiFi
& 64.62
& \textbf{67.36}
& \textcolor{green!45!black}{+2.74}
& 77.27
& \textbf{83.59}
& \textcolor{green!45!black}{+6.32} \\

RFID
& 41.60
& \textbf{42.54}
& \textcolor{green!45!black}{+0.94}
& 44.23
& \textbf{49.18}
& \textcolor{green!45!black}{+4.95} \\

mmWave+WiFi
& 89.92
& \textbf{90.74}
& \textcolor{green!45!black}{+0.82}
& 91.86
& \textbf{94.38}
& \textcolor{green!45!black}{+2.52} \\

mmWave+RFID
& 83.76
& \textbf{85.84}
& \textcolor{green!45!black}{+2.08}
& 89.91
& \textbf{90.58}
& \textcolor{green!45!black}{+0.67} \\

WiFi+RFID
& 67.49
& \textbf{67.95}
& \textcolor{green!45!black}{+0.46}
& 74.55
& \textbf{80.82}
& \textcolor{green!45!black}{+6.27} \\

All
& 90.09
& \textbf{91.01}
& \textcolor{green!45!black}{+0.92}
& 90.45
& \textbf{95.00}
& \textcolor{green!45!black}{+4.55} \\

\midrule

\textbf{Avg.}
& 74.26
& \textbf{75.57}
& \textcolor{green!45!black}{+1.31}
& 79.55
& \textbf{83.44}
& \textcolor{green!45!black}{+3.89} \\

\bottomrule
\end{tabular*}

\vspace{-1mm}
\end{table*}

\subsection{Geometry Diagnostics and Visualizations}
\label{sec:exp_geom}

We next examine whether the performance gains of \regName are accompanied by systematic changes in representation geometry.

\subsubsection{Diagnostics on Modality-Specific Embeddings}
\label{sec:embedding_diagnostics}

We characterize the learned representation geometry from three complementary perspectives: semantic organization, representation diversity, and cross-modal agreement. Semantic organization is evaluated using the between-to-within-class distance ratio (B/W) and $k$-nearest-neighbor (kNN) purity, which quantify class separation and local semantic consistency, respectively. Representation
diversity is assessed through spectral statistics of the modality-specific embeddings, capturing how broadly information is distributed across feature
directions. Cross-modal agreement is characterized by paired drift together with tolerance-based diagnostics, which quantify both the frequency and
magnitude of deviations beyond the prescribed agreement region. Detailed definitions of all geometry diagnostics are provided in Appendix~\ref{app:geometry_diagnostics}.

\begin{table*}[t]
\centering
\footnotesize
\setlength{\tabcolsep}{5pt}
\renewcommand{\arraystretch}{1.08}
\caption{
\textbf{Representation-geometry diagnostics.}
DGL is used as the base method on CREMA-D and AGM on UR-FUNNY.
Higher B/W, kNN purity, $r_{\mathrm{eff}}$, and task performance are better;
lower drift, band-violation rate $\rho_\tau$, and excess violation $E_\tau$
are better.
}
\label{tab:direct_geometry}
\vspace{1mm}
\begin{tabular}{llcccccc}
\toprule
\textbf{Data}
& \textbf{Method}
& \textbf{B/W}$\uparrow$
& \textbf{kNN}$\uparrow$
& $r_{\mathrm{eff}}\uparrow$
& \textbf{Drift}
& $\rho_{\tau}/E_{\tau}\downarrow$
& \textbf{Acc./F1 (Single run)}$\uparrow$ \\
\midrule
CREMA-D
& DGL
& 1.38
& 0.51
& 4.09
& 0.51
& 0.28 / 0.0013
& 77.70 \\
CREMA-D
& DGL + \regName
& \textbf{1.43}
& \textbf{0.52}
& \textbf{4.18}
& \textbf{0.49}
& \textbf{0.14 / 0.00027}
& \textbf{79.26} \\
\midrule
UR-FUNNY
& AGM
& 1.01
& 0.53
& 12.58
& 1.42
& 1.00 / 1.14
& 63.68 \\
UR-FUNNY
& AGM + \regName
& 1.01
& \textbf{0.54}
& \textbf{14.20}
& \textbf{0.49}
& \textbf{0.90 / 0.02}
& \textbf{68.31} \\
\bottomrule
\end{tabular}
\vspace{-2mm}
\end{table*}

Table~\ref{tab:direct_geometry} shows a consistent geometric effect across
datasets. On CREMA-D, adding \regName to DGL improves class separation and
neighborhood purity, increases effective rank, and reduces both paired drift
and agreement violations. On UR-FUNNY, adding \regName to AGM preserves or
improves semantic structure while increasing effective rank from $12.58$ to
$14.20$ and substantially reducing paired drift from $1.42$ to $0.49$.
The excess agreement violation also drops from $1.14$ to $0.02$.

These results suggest that \regName does not simply spread representations
indiscriminately. Instead, modality-wise diversity broadens the representation
space while bounded agreement prevents excessive cross-modal drift, yielding
a geometry that preserves semantic structure while improving downstream
performance.

\subsubsection{Qualitative Representation Geometry}
\label{sec:tsne}

\begin{figure*}[t]
    \centering
    \includegraphics[width=0.88\textwidth]{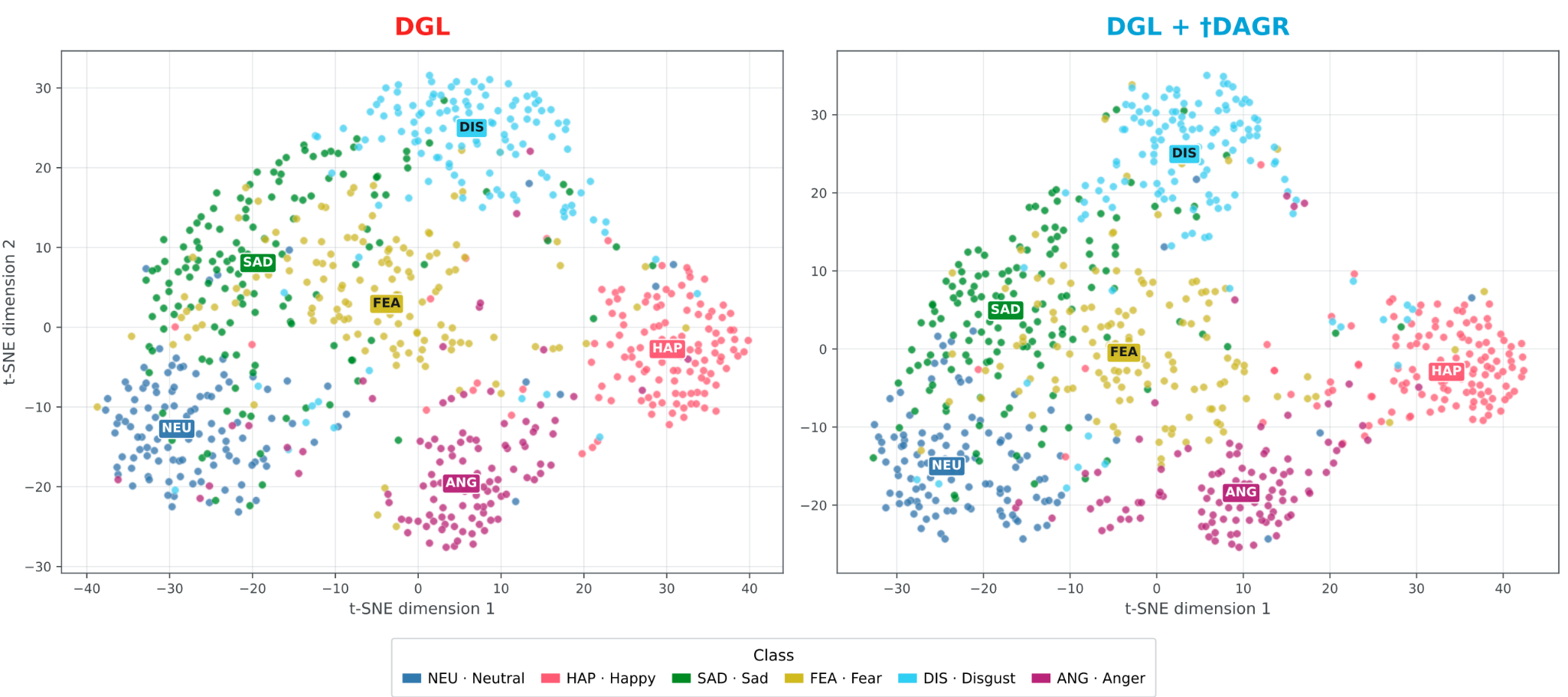}
    \caption{
    \textbf{t-SNE visualization of class-wise representations on CREMA-D.}
    We compare DGL (left) with DGL+\regName (right), using the same baseline
    as in Table~\ref{tab:direct_geometry}. Colors denote emotion classes.
    }
    \label{fig:tsne}
\end{figure*}

We further visualize the learned representation geometry on CREMA-D using t-SNE. As shown in Figure~\ref{fig:tsne}, after introducing \regName, the representations exhibit more coherent class structure, with several emotion categories becoming more compact and better separated. This qualitative trend is consistent with the improved class separation, neighborhood purity, and effective rank reported
in Table~\ref{tab:direct_geometry}, providing complementary evidence that \regName promotes a more structured representation geometry.

\subsubsection{Qualitative Cross-Modal Agreement}
\label{sec:qual_crossmodal}

\begin{figure*}[t]
    \centering
    \includegraphics[width=0.8\textwidth]{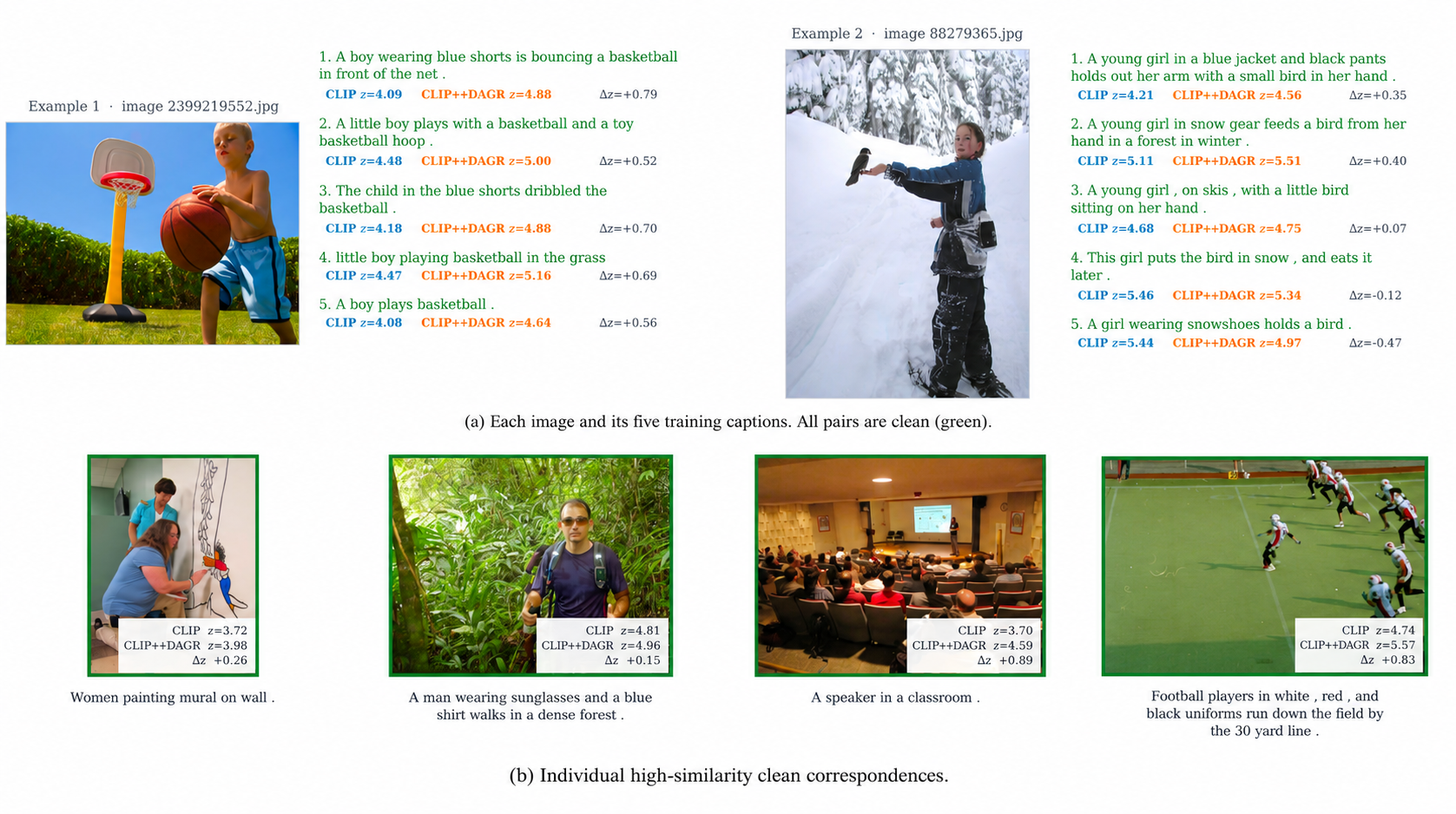}
    \caption{
    \textbf{Qualitative analysis of image--text agreement.}
    \textbf{(a)} Clean image--caption pairs with similarity changes after adding \regName.
    \textbf{(b)} Representative high-similarity clean correspondences.
    }
    \label{fig:qual_crossmodal}
\end{figure*}

In addition to the global representation diagnostics, we qualitatively inspect how \regName affects individual image--text correspondences. Figure~\ref{fig:qual_crossmodal} compares standardized image--text similarity scores produced by the original CLIP model and CLIP+\regName. For each image--caption pair, we report the similarity $z$-score, defined by standardizing its cosine similarity within the corresponding similarity distribution, together with $\Delta z$.

The examples illustrate two aspects of bounded agreement. First, for many semantically consistent image--caption pairs, \regName substantially strengthens cross-modal correspondence. For instance, several captions describing the basketball example obtain increases of approximately $0.5$--$0.8$ in standardized similarity, while the high-similarity pairs in Fig.~\ref{fig:qual_crossmodal}(b) also consistently receive higher scores. Second, the effect is not a uniform increase in similarity: some already well-matched captions exhibit little change or even a moderate decrease. This behavior is consistent with the bounded-agreement objective, which does not continuously maximize cross-modal similarity once sufficient agreement has been achieved.

.

\subsection{Cross-Framework Generalization and Robustness}
\subsubsection{Plug-in generality}
\begin{figure*}[t]
    \centering
    \includegraphics[width=0.8\textwidth]{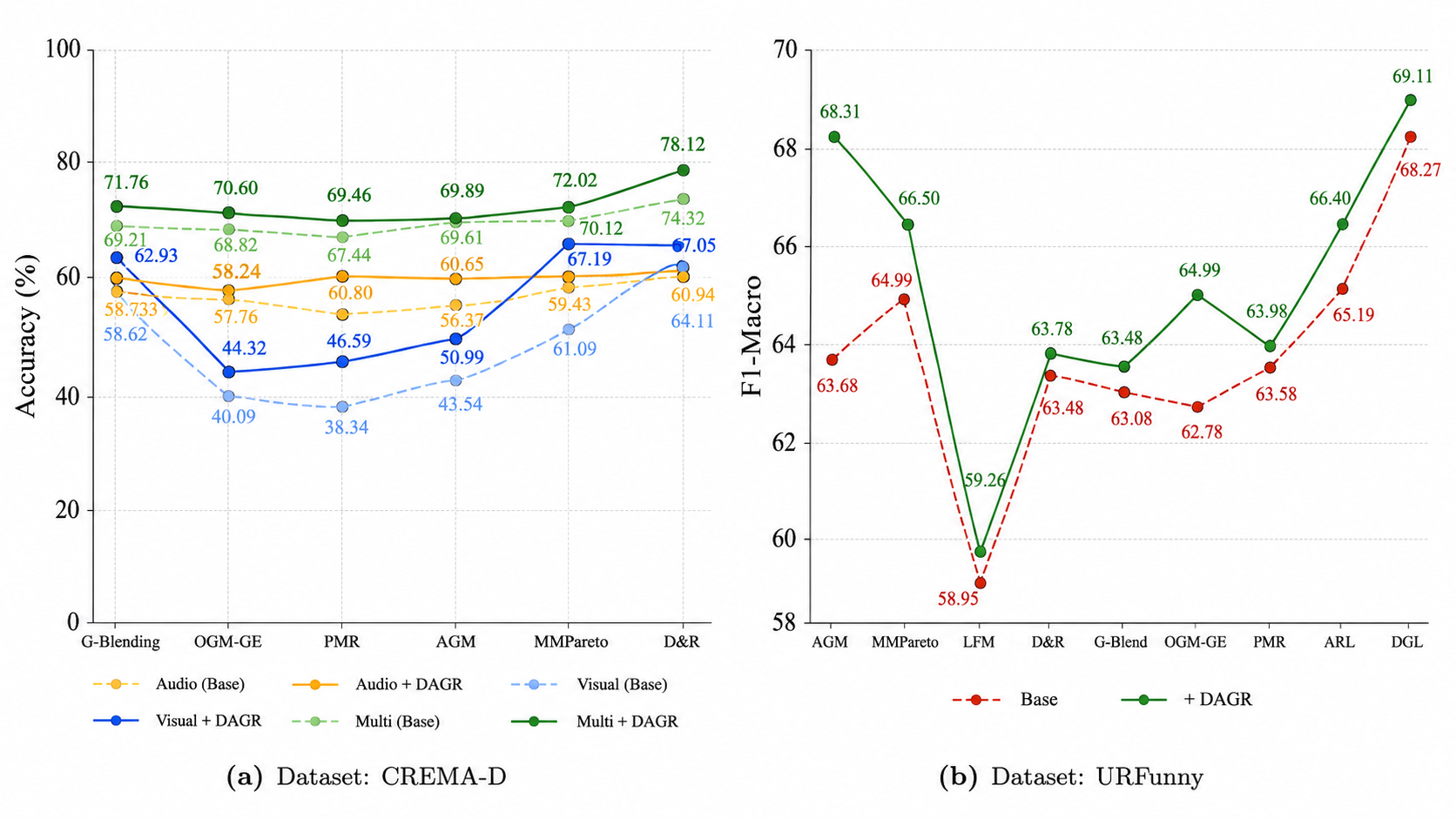}
    \caption{
    \textbf{Plug-in generality of \regName across datasets and multimodal
    optimization methods.}
    \textbf{(a)} CREMA-D results for audio, visual, and multimodal evaluation
    across six representative backbones.
    Dashed/light curves denote the original backbone performance, while
    solid/darker curves denote the corresponding results after adding
    \regName.
    \textbf{(b)} F1-Macro results on the tri-modal URFunny benchmark across
    nine representative methods, comparing each original backbone with its
    \regName-augmented counterpart.
    Numbers indicate the corresponding absolute performance.
    }
    \label{fig:plugin_generality}
\end{figure*}

To examine whether \regName is tied to a particular multimodal optimizer or
dataset, we further apply it as a plug-in regularizer to a diverse set of
existing multimodal learning methods.
Figure~\ref{fig:plugin_generality}(a) reports results on CREMA-D for six
representative backbones---G-Blending, OGM-GE, PMR, AGM, MMPareto, and
D\&R---under audio-only, visual-only, and multimodal evaluation.
Across the resulting $6\times3$ backbone--modality settings, \regName improves
17 out of 18 cases.
In particular, the visual and multimodal results improve for all six
backbones, while the audio results improve in five cases and remain nearly
unchanged for D\&R ($-0.17$ points).

We observe the same trend on the tri-modal URFunny benchmark in
Figure~\ref{fig:plugin_generality}(b).
Across nine representative methods, adding \regName consistently improves F1-Macro, with gains ranging from $+0.30$ to $+4.63$ points.
The improvements are especially pronounced for AGM ($+4.63$) and OGM-GE
($+2.21$), while even stronger baselines such as ARL and DGL retain positive gains. Together, these results indicate that \regName is complementary to a broad range of multimodal optimization strategies and generalizes across both datasets and modality configurations, rather than being specific to DGL or to a particular training pipeline.

\subsubsection{Robustness to Training-Time Modality Imbalance}
\label{sec:robustness_main}

We further examine whether \regName remains effective when one modality is
under-represented during training. On CREMA-D, we randomly retain only a
fraction of either the audio or visual observations, while keeping the other
modality and class labels available for all training samples. Evaluation is
performed on the fully observed test set, allowing us to assess whether
\regName preserves useful representations under asymmetric modality exposure.

\begin{figure*}[t]
    \centering
    \includegraphics[width=0.8\textwidth]{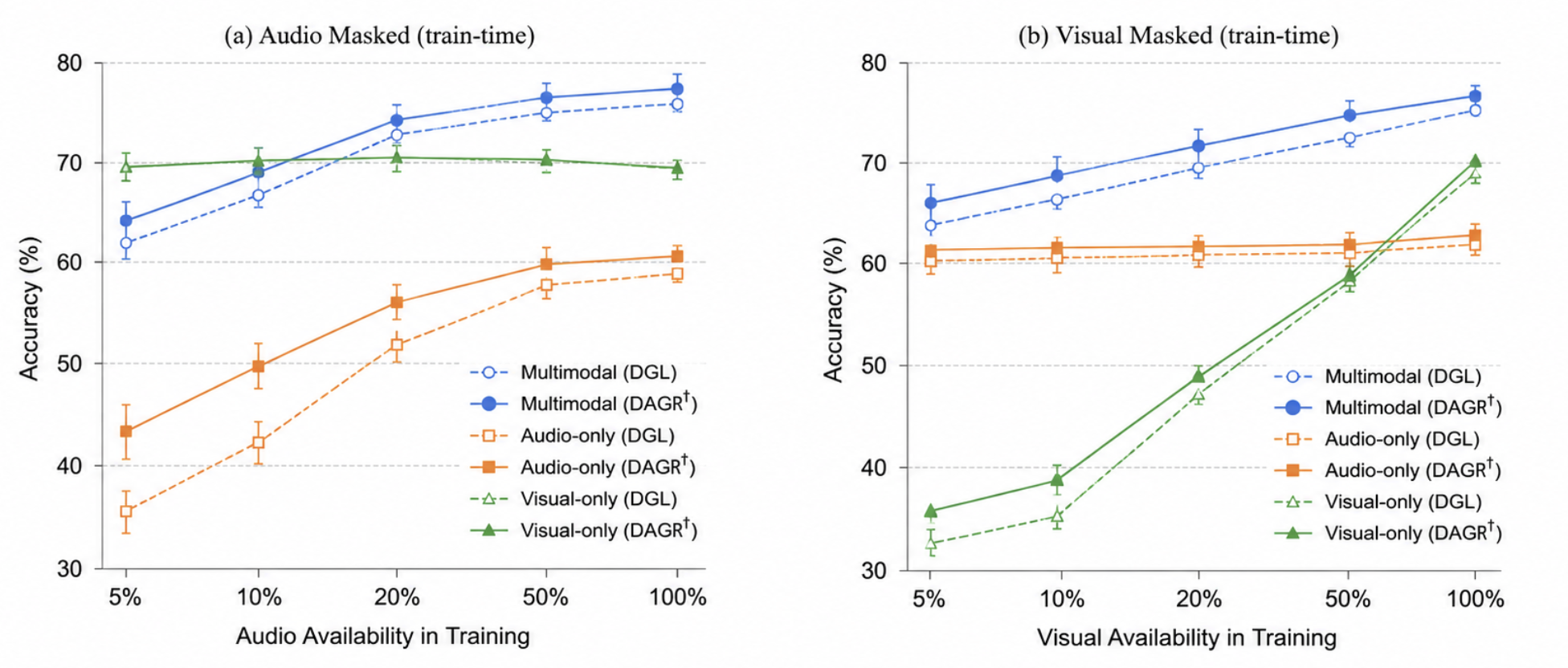}
    \vspace{-2mm}
    \caption{\textbf{Robustness to training-time modality imbalance on CREMA-D.} Dashed lines denote DGL and solid lines denote DGL+\regName under (a) audio masking and (b) visual masking. Error bars show standard deviations over three seeds.
    }
    \label{fig:modality_masking}
\end{figure*}

Figure~\ref{fig:modality_masking}(a) shows that under audio masking,
\regName is particularly effective at moderate imbalance. At $50\%$ and
$20\%$ audio availability, multimodal accuracy improves by $+0.8$ and
$+2.2$ points, respectively, while audio-only accuracy improves by $+2.1$
and $+4.1$ points. The gain is largest at $20\%$ availability, where the
audio modality provides substantially less training evidence but still
remains informative. Under more extreme scarcity ($10\%$ and $5\%$),
the effect becomes mixed, indicating that geometric regularization cannot
fully compensate when modality-specific supervision becomes too sparse.

A complementary pattern appears under visual masking
(Figure~\ref{fig:modality_masking}(b)). Multimodal performance remains
largely preserved across all availability levels, with gains of $+0.5$
points at both $50\%$ and $5\%$ availability. Moreover, under severe visual
scarcity, the visual-only branch improves by $+0.4$, $+0.9$, and $+1.0$
points at $20\%$, $10\%$, and $5\%$ availability, respectively.
Together, these results suggest that \regName does not simply favor one
modality, but can help preserve modality-specific structure when cross-modal
training evidence becomes imbalanced.

.

\subsubsection{Robustness to Noisy Cross-Modal Correspondence}
\label{sec:noisy_correspondence}

We further test whether bounded agreement remains effective when paired
modalities become unreliable. Following a controlled corruption protocol,
we replace an increasing fraction of image--text correspondences with noisy
pairs during training and evaluate retrieval on the clean test sets of
Flickr30K and COCO.

\begin{figure}[t]
    \centering
    \includegraphics[width=0.7\linewidth]{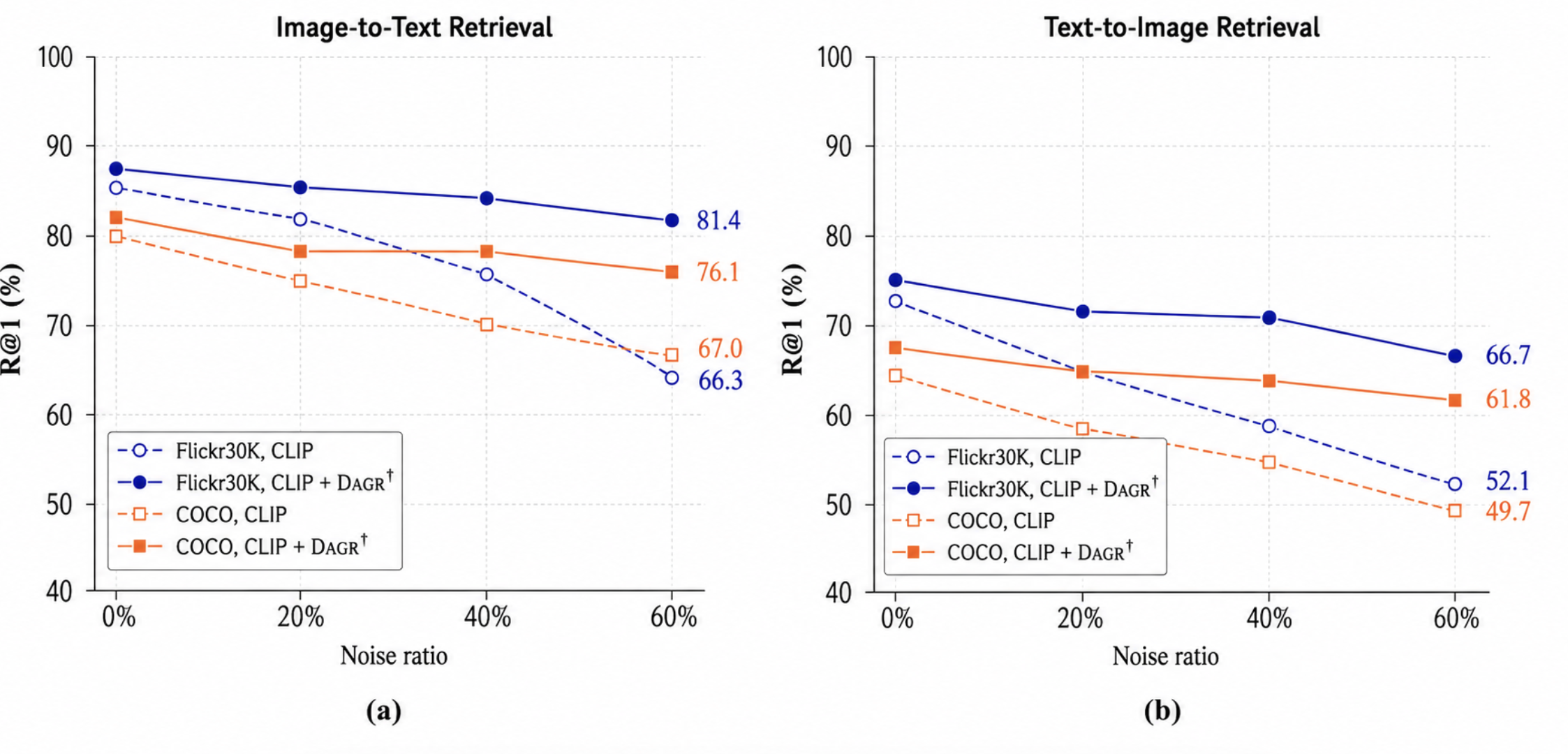}
    \vspace{-2mm}
    \caption{
    \textbf{Robustness to noisy image--text correspondences.}
    R@1 performance under increasing correspondence noise on Flickr30K and
    COCO for (a) image-to-text and (b) text-to-image retrieval.
    Solid lines denote CLIP+\regName and dashed lines denote CLIP.
    }
    \label{fig:noisy_retrieval}
    \vspace{-2mm}
\end{figure}

As shown in Figure~\ref{fig:noisy_retrieval}, \regName consistently improves
retrieval robustness, with the advantage becoming substantially larger as
correspondence noise increases. At $60\%$ noise, the R@1 gains reach
$+15.1$ and $+9.1$ points for image-to-text retrieval on Flickr30K and
COCO, and $+14.6$ and $+12.1$ points for text-to-image retrieval,
respectively. In contrast to always-on alignment, which can directly
propagate erroneous pairwise supervision, bounded agreement limits unnecessary
cross-modal attraction once representations enter the tolerance region.
The widening performance gap under stronger corruption therefore provides
additional evidence that \regName is particularly beneficial when exact
cross-modal correspondence is unreliable. Full R@1/R@5/R@10 results are
reported in Appendix~\ref{sec:robustness:noisy_image_text}.

\subsection{Ablation studies}
\subsubsection{Effect of the Agreement Radius $\tau$}
\label{sec:ablation_tau}

\begin{figure}[t]
    \centering
    \includegraphics[width=0.7\linewidth]{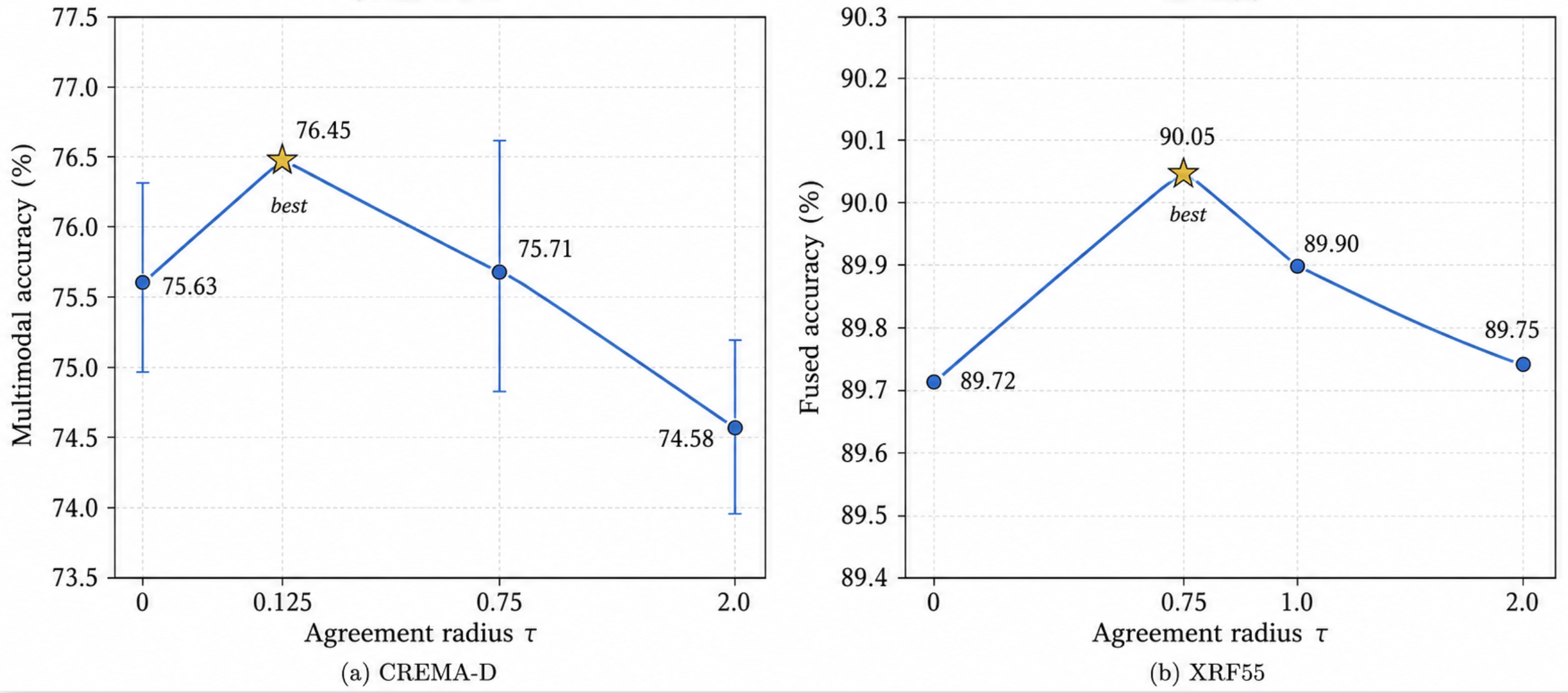}
    \caption{
    \textbf{Effect of the agreement radius $\tau$.}
    Multimodal accuracy on CREMA-D and fused accuracy on XRF55 under different
    agreement radiius. This controlled sweep uses a fixed-coefficient ablation configuration and is therefore not directly comparable to the Pareto-balanced main results.
    }
    \label{fig:tau_sweep}
\end{figure}

We investigate the effect of the agreement radius $\tau$, which controls the tolerance of cross-modal discrepancy in \regName. As shown in Figure~\ref{fig:tau_sweep}, exact agreement ($\tau=0$) is suboptimal on both CREMA-D and XRF55. On CREMA-D, the multimodal accuracy increases from $75.63\%$ at $\tau=0$ to $76.45\%$ at $\tau=0.125$, but decreases to $74.58\%$ at $\tau=2.0$. Similarly, XRF55 peaks at $90.05\%$ with $\tau=0.75$, compared with $89.72\%$ at $\tau=0$ and $89.75\%$ at $\tau=2.0$.

These results support the bounded-agreement design of \regName: overly strict alignment can suppress modality-specific variation, whereas an excessively large radius weakens cross-modal coupling. A moderate positive $\tau$ therefore provides a better balance between cross-modal consistency and modality-specific flexibility.

\subsubsection{Comparison with Alternative Geometric Objectives}
\label{sec:ablation_controls}

We further compare \regName with alternative geometric regularizers under the
same DGL backbone and training protocol.
\emph{Naive Align} continuously minimizes paired cross-modal distance,
\emph{VICReg-style} applies invariance, variance, and covariance regularization,
\emph{Barlow Twins-style} uses cross-correlation decorrelation,
\emph{Unit-sphere only} keeps only $\ell_2$ normalization, and
\emph{InfoNCE-style Align} replaces bounded agreement with contrastive
pairwise alignment.

\begin{table}[t]
\centering
\small
\setlength{\tabcolsep}{6pt}
\renewcommand{\arraystretch}{1.15}
\vspace{2mm}
\caption{
\textbf{Comparison with alternative geometric objectives on CREMA-D.}
All variants use the same DGL backbone and training protocol.
}
\vspace{2mm}
\label{tab:cremad_extra_baselines}
\begin{tabular}{l|ccc}
\toprule
\textbf{Method}
& Audio & Visual & Multi \\
\midrule
Naive Align
& 60.65$\pm{0.86}$
& \best{72.21}$\pm{1.07}$
& 77.98$\pm{1.11}$ \\

VICReg-style
& 59.18$\pm{1.67}$
& 71.35$\pm{1.00}$
& \second{78.12}$\pm{0.94}$ \\

Barlow Twins-style
& 59.09$\pm{0.62}$
& 61.32$\pm{3.41}$
& 71.07$\pm{0.81}$ \\

Unit-sphere only
& 60.65$\pm{1.30}$
& 69.84$\pm{0.22}$
& 75.81$\pm{1.52}$ \\

InfoNCE-style Align
& \best{63.31}$\pm{0.03}$
& 70.17$\pm{1.16}$
& 76.14$\pm{0.65}$ \\

\midrule
\regName
& \second{62.98}$\pm{0.50}$
& \second{72.10}$\pm{0.55}$
& \best{79.40}$\pm{0.14}$ \\
\bottomrule
\end{tabular}
\end{table}

As shown in Table~\ref{tab:cremad_extra_baselines}, \regName achieves the best multimodal accuracy while remaining strong on both unimodal branches. Unit-sphere normalization and InfoNCE-style alignment are clearly weaker in multimodal performance, while Naive Align improves the visual branch at the expense of audio accuracy. VICReg-style regularization is competitive but also weakens the audio branch, and Barlow Twins-style decorrelation performs substantially worse.

Together with the $\tau$-sweep, these results indicate that the gain of \regName is not explained by normalization or generic alignment alone. Its bounded-agreement objective better balances cross-modal consistency with the preservation of modality-specific information.

\subsubsection{Component-wise ablation}
We conduct a component-wise ablation study to examine the roles of the two operational terms in \regName: modality-wise dispersion $\mathcal{L}_{\mathrm{d}}$ and agreement-band anchoring $\mathcal{L}_{\mathrm{a}}$.
We compare the baseline training pipeline without geometry-aware regularization, applying only $\mathcal{L}_{\mathrm{d}}$, applying only $\mathcal{L}_{\mathrm{a}}$, and using the full \regName objective.

As shown in Table~\ref{tab:ablation_component}, both terms improve over the baseline when applied individually, but they emphasize different aspects of the feasible geometry.
Dispersion tends to improve the audio branch, while anchoring yields larger gains for the visual branch in several settings.
The full objective achieves the best multimodal accuracy on both CREMA-D and Kinetics-Sounds.
This ablation should be interpreted as a mechanism check: the two terms are not independent novelty claims, but complementary operational components that instantiate diversity under bounded agreement.

\begin{table*}[t]
\centering
\footnotesize
\setlength{\tabcolsep}{4.8pt}
\renewcommand{\arraystretch}{1.08}

\caption{
Component-wise ablation of geometry-aware regularization on CREMA-D and Kinetics-Sounds.
We report unimodal (Audio, Visual) and multimodal (Multi) classification accuracy (\%).
}
\label{tab:ablation_component}
\vspace{2mm}
\begin{tabular}{cc|ccc|ccc}
\toprule
\multicolumn{2}{c|}{Component}
& \multicolumn{3}{c|}{CREMA-D}
& \multicolumn{3}{c}{Kinetics-Sounds} \\
\cmidrule(lr){1-2} \cmidrule(lr){3-5} \cmidrule(lr){6-8}
Disp & Anchor
& Audio & Visual & Multi
& Audio & Visual & Multi \\
\midrule
            &
& 62.17$\pm{0.79}$ & 70.31$\pm{0.79}$ & 77.65$\pm{0.79}$
& 50.63$\pm0.54$ & 59.83$\pm0.63$ & 75.44$\pm0.49$ \\
\checkmark &
& \best{63.26}$\pm{0.59}$ & 71.76$\pm{1.00}$ & 77.81$\pm{1.52}$
& \second{52.52}$\pm{0.84}$ & 62.62$\pm{0.23}$ & \second{77.04}$\pm{0.43}$ \\
            & \checkmark
& 62.55$\pm{0.73}$ & \best{72.92}$\pm{0.78}$ & \second{77.98}$\pm{1.11}$
& 51.67$\pm{0.41}$ & \second{63.04}$\pm{0.88}$ & 76.62$\pm{0.17}$ \\
\midrule
\checkmark & \checkmark
& \second{62.98}$\pm{0.50}$ & \second{72.10}$\pm{0.55}$ & \best{79.40}$\pm{0.14}$
& \best{52.79}$\pm{0.59}$ & \best{63.25}$\pm{0.47}$ & \best{77.15}$\pm{0.58}$ \\
\bottomrule
\end{tabular}
\end{table*}

\section{Conclusions, Limitations and Future Work}

We introduced \regName, a lightweight geometric regularizer that promotes modality-wise diversity under bounded cross-modal agreement. Across diverse multimodal benchmarks and tasks, \regName consistently improves performance without architectural changes or inference-time overhead, while also yielding more favorable representation geometry and robustness.

Our evaluation mainly focuses on medium-scale discriminative settings. Extending \regName to large-scale multimodal foundation models, generative tasks, and more complex temporal settings is an important direction for future work.

\bibliography{tmlr}
\bibliographystyle{tmlr}

\newpage
\appendix

\section*{Appendix}
\addcontentsline{toc}{section}{Appendix Contents}

\begingroup
  \setcounter{tocdepth}{2}
  \startcontents
  \printcontents{}{1}{}
\endgroup

\section{Formal Analysis of \regName}
\label{appendix:proof}

This appendix provides the formal statements that support the geometric interpretation of \regName. 
Throughout the analysis, we write $z$ for the $\ell_2$-normalized embedding $\tilde z$. 
The results are intended to isolate the representation-level bias induced by the proposed regularizer. 
They do not claim to fully characterize the non-convex optimization dynamics of deep multimodal networks.

\subsection{An Illustrative Non-identifiability Example under a Linear Head}
\label{app:geom_ambiguity}

We first give a restricted example showing why optimizing a supervised task
loss need not determine the geometry of an intermediate representation.  The
statement concerns raw, pre-normalization representations and a linear task
head; it is intended as an illustration rather than a characterization of
arbitrary nonlinear predictors.

\begin{proposition}[Geometric non-identifiability under a linear head]
\label{prop:app_geom_ambiguity}
Let $z^m\in\mathbb{R}^{d}$ be a modality-specific representation and let its
logits be $W_m z^m+b_m$, where $W_m\in\mathbb{R}^{K\times d}$.  If a linear map
$T_m\in\mathbb{R}^{d\times d}$ satisfies
\[
W_mT_m=W_m,
\]
then replacing every $z^m$ by $T_mz^m$ leaves the logits, and hence any loss
that depends on $z^m$ only through these logits, unchanged.  Nevertheless,
pairwise distances and second-order geometry can change, since
\[
\|T_m(z_i^m-z_j^m)\|_2^2
=(z_i^m-z_j^m)^\top T_m^\top T_m(z_i^m-z_j^m)
\]
and
\[
\operatorname{Cov}(T_mz^m)
=T_m\operatorname{Cov}(z^m)T_m^\top.
\]
If $\ker(W_m)\neq\{0\}$, infinitely many such transformations exist.
\end{proposition}

\begin{proof}
The logit invariance follows directly from
\[
W_m(T_mz^m)+b_m=W_mz^m+b_m.
\]
If $\ker(W_m)\neq\{0\}$, take $T_m=I+A_m$ for any nonzero map $A_m$ whose
image lies in $\ker(W_m)$.  Then $W_mA_m=0$ and therefore $W_mT_m=W_m$.
The displayed distance and covariance identities show that such a map need
not preserve representation geometry.
\end{proof}

This example only establishes non-identifiability under a linear head.  It
motivates explicit geometric control, but does not by itself imply that a
particular geometry improves downstream performance.

\subsection{A Population-Level Interpretation of the Dispersion Energy}
\label{app:me}

Let $\nu$ be the normalized surface measure on the unit sphere
$\mathbb{S}^{d-1}$, and suppose that the normalized representation of modality
$m$ has a sufficiently regular density $p_m$ with respect to $\nu$.  For the
RBF kernel
\[
k_t(z,z')=\exp\!\left(-t\|z-z'\|_2^2\right),
\]
define its population self-interaction energy by
\[
\mathcal E_t(p_m)
=\iint_{\mathbb S^{d-1}\times\mathbb S^{d-1}}
k_t(z,z')p_m(z)p_m(z')\,d\nu(z)\,d\nu(z').
\]
Because the sphere and the kernel are rotationally invariant,
\[
c_{d,t}
=\int_{\mathbb S^{d-1}}k_t(z,z')\,d\nu(z')
\]
does not depend on $z$.  After normalization by $c_{d,t}$, $k_t$ is an
approximate identity as $t\to\infty$.  Consequently,
\[
\mathcal E_t(p_m)
=c_{d,t}
\left(
\int_{\mathbb S^{d-1}}p_m(z)^2\,d\nu(z)+o(1)
\right),
\qquad t\to\infty.
\]
Writing the R\'enyi-$2$ entropy relative to $\nu$ as
\[
H_{2,\nu}(p_m)
=-\log\int_{\mathbb S^{d-1}}p_m(z)^2\,d\nu(z),
\]
we obtain
\[
\log\mathcal E_t(p_m)
=\log c_{d,t}-H_{2,\nu}(p_m)+o(1).
\]
Thus, in the narrow-kernel regime, minimizing the log self-interaction energy
corresponds up to a distribution-independent constant to maximizing
modality-wise R\'enyi-$2$ entropy.

This interpretation can be combined with bounded cross-modal agreement.  For
paired normalized representations $Z_\theta^1,\ldots,Z_\theta^M$, define
\[
\mathcal A_\tau(\theta)
=\mathbb E\!\left[
\frac{1}{M(M-1)}
\sum_{m\neq n}
\bigl(\|Z_\theta^m-Z_\theta^n\|_2-\tau\bigr)_+^2
\right].
\]
A population-level constrained objective is
\[
\min_\theta
\left\{
\mathcal L_{\mathrm{task}}(\theta)
-\frac{\lambda_{\mathrm d}}{M}
\sum_{m=1}^{M}H_{2,\nu}(p_m)
\right\}
\quad\mathrm{s.t.}\quad
\mathcal A_\tau(\theta)\leq\epsilon.
\]
Its penalized form replaces the constraint by
$\lambda_{\mathrm a}\mathcal A_\tau(\theta)$.  Replacing each negative entropy
by $\log\mathcal E_t(p_m)$ gives the population analogue of the geometric
regularizer used by \regName.

For a mini-batch $\{z_i^m\}_{i=1}^{B}$, the implementation uses
\[
\widehat{\mathcal E}_{t,m}
=\frac{1}{B(B-1)}
\sum_{i\neq j}
\exp\!\left(-t\|z_i^m-z_j^m\|_2^2\right),
\qquad
\mathcal L_{\mathrm d}^m=\log\widehat{\mathcal E}_{t,m}.
\]
The U-statistic $\widehat{\mathcal E}_{t,m}$ is unbiased for
$\mathcal E_t(p_m)$, but its logarithm is not an unbiased estimator of
$\log\mathcal E_t(p_m)$.  Hence the mini-batch loss is a stochastic surrogate,
and the entropy equivalence is an asymptotic population interpretation rather
than an exact finite-batch identity.  If $p_m$ is not absolutely continuous
with respect to the sphere measure, the kernel energy remains well defined,
but the stated density-based entropy interpretation is not literal.

\subsection{A Small-$t$ Expansion of the RBF Energy}
\label{app:dispersion_spectrum}

The preceding entropy interpretation concerns the narrow-kernel limit
$t\to\infty$.  A separate expansion around $t=0$ relates the same kernel
energy locally to the representation mean and covariance.

\begin{theorem}[Small-$t$ expansion]
\label{thm:dispersion_spectrum}
Let $z,z'$ be independent copies of a random vector on
$\mathbb S^{d-1}$.  Define
\[
\mu=\mathbb E[z],
\qquad
\Sigma=\operatorname{Cov}(z),
\qquad
S=\mathbb E[zz^\top]=\Sigma+\mu\mu^\top.
\]
Then, as $t\to0$,
\[
\log\mathbb E\exp\!\left(-t\|z-z'\|_2^2\right)
=-2t\bigl(1-\|\mu\|_2^2\bigr)
+2t^2
\left(
\operatorname{tr}(\Sigma^2)+2\mu^\top\Sigma\mu
\right)
+O(t^3).
\]
In particular, if the representation is centered, then
$\operatorname{tr}(\Sigma)=1$ and
\[
\log\mathbb E\exp\!\left(-t\|z-z'\|_2^2\right)
=-2t+2t^2\operatorname{tr}(\Sigma^2)+O(t^3).
\]
Therefore, to second order in the centered small-$t$ regime, minimizing the
RBF energy discourages spectral concentration and increases the
participation-ratio rank
\[
\operatorname{PR}(\Sigma)
=\frac{\operatorname{tr}(\Sigma)^2}
{\operatorname{tr}(\Sigma^2)}
=\frac{1}{\operatorname{tr}(\Sigma^2)}.
\]
\end{theorem}

\begin{proof}
Let $u=z^\top z'$.  Since $z$ and $z'$ are normalized,
\[
\log\mathbb E\exp\!\left(-t\|z-z'\|_2^2\right)
=-2t+\log\mathbb E\exp(2tu).
\]
Because $u\in[-1,1]$, the cumulant expansion is valid around $t=0$ and gives
\[
\log\mathbb E\exp(2tu)
=2t\mathbb E[u]+2t^2\operatorname{Var}(u)+O(t^3).
\]
Independence implies
\[
\mathbb E[u]=\|\mu\|_2^2,
\qquad
\mathbb E[u^2]=\operatorname{tr}(S^2).
\]
Moreover,
\[
\begin{aligned}
\operatorname{Var}(u)
&=\operatorname{tr}(S^2)-\|\mu\|_2^4 \\
&=\operatorname{tr}(\Sigma^2)+2\mu^\top\Sigma\mu.
\end{aligned}
\]
Substitution proves the general expansion.  If $\mu=0$, then
$S=\Sigma$ and
$\operatorname{tr}(\Sigma)=\mathbb E\|z\|_2^2=1$, which yields the centered
special case and the stated participation-ratio identity.
\end{proof}

This result is deliberately local: without centering, the first-order mean
term and the covariance--mean interaction are also present, and at the finite
$t$ used in training higher-order terms need not be negligible.  The theorem
therefore motivates the participation-ratio rank of the centered covariance
as a diagnostic; it does not imply monotonic improvement of the Shannon
effective rank for arbitrary $t$.  The $t\to0$ expansion here and the
$t\to\infty$ entropy interpretation in Appendix~\ref{app:me} are complementary
asymptotic views, not simultaneous exact descriptions of a finite choice of
$t$.

\subsection{Exact Geometry and Gradients of \regName}
\label{app:exact_geometry}

We finally record the exact finite-sample geometry of the two terms.  These
identities do not require either asymptotic regime above.

\paragraph{Dispersion gradient.}
For one modality, let
\[
S_m
=\sum_{p\neq q}
\exp\!\left(-t\|z_p^m-z_q^m\|_2^2\right),
\qquad
\mathcal L_{\mathrm d}^m
=\log\frac{S_m}{B(B-1)}.
\]
Because the sum is over ordered pairs, its ambient Euclidean gradient is
\[
\nabla_{z_i^m}\mathcal L_{\mathrm d}^m
=-\frac{4t}{S_m}
\sum_{j\neq i}
\exp\!\left(-t\|z_i^m-z_j^m\|_2^2\right)
(z_i^m-z_j^m).
\]
Hence the negative ambient gradient is a weighted repulsive force, with the
largest kernel weights assigned to nearby representations.

If $z_i^m$ is obtained by normalizing a raw representation $h_i^m\neq0$,
\[
z_i^m=\frac{h_i^m}{\|h_i^m\|_2},
\]
then the normalization Jacobian gives
\[
\nabla_{h_i^m}\mathcal L
=\frac{1}{\|h_i^m\|_2}
\left(I-z_i^m(z_i^m)^\top\right)
\nabla_{z_i^m}\mathcal L.
\]
Thus the gradient applied to the raw representation is the scaled tangent-space
projection of the ambient force.

\paragraph{Bounded-agreement geometry.}
For one normalized cross-modal pair $x,y\in\mathbb S^{d-1}$, write
\[
r=\|x-y\|_2,
\qquad
\ell_\tau(x,y)=(r-\tau)_+^2.
\]
Let $\mathbb B_2(\tau)=\{v\in\mathbb R^d:\|v\|_2\leq\tau\}$.  Then
\[
\ell_\tau(x,y)
=\operatorname{dist}^2\!\left(x-y,\mathbb B_2(\tau)\right).
\]
Thus the loss penalizes only the component of the cross-modal difference that
lies outside the radius-$\tau$ ball.  Equivalently, for $0\leq\tau\leq2$ and
$\vartheta=\arccos(x^\top y)$,
\[
r^2=2-2x^\top y=4\sin^2\!\left(\frac{\vartheta}{2}\right),
\]
so that
\[
r\leq\tau
\quad\Longleftrightarrow\quad
x^\top y\geq1-\frac{\tau^2}{2}
\quad\Longleftrightarrow\quad
\vartheta\leq2\arcsin\!\left(\frac{\tau}{2}\right).
\]

The vector gradients are
\[
\nabla_x\ell_\tau(x,y)
=
\begin{cases}
2\left(1-\dfrac{\tau}{r}\right)(x-y), & r>\tau,\\[6pt]
0, & r\leq\tau,
\end{cases}
\qquad
\nabla_y\ell_\tau(x,y)=-\nabla_x\ell_\tau(x,y).
\]
The squared hinge is differentiable at $r=\tau$, where its gradient is zero.
Consequently, bounded agreement has an exact zero-gradient tolerance region.
For $\tau=0$, it reduces to squared-distance alignment; for $\tau\geq2$, it is
identically zero on normalized representations.  Importantly, decreasing the
weight of an always-on squared-alignment loss rescales its gradient but does
not change its support, whereas changing $\tau$ changes which pairs receive an
agreement gradient.

\section{Implementation and Experiment Details}
\label{app:implementation}

\subsection{Datasets Details}
\label{sec:datasets}

We evaluate \regName on seven multimodal benchmark datasets spanning
audio--visual recognition, three-modality language--audio--visual learning,
image--text clustering and retrieval, and RF-based human activity recognition.

\textbf{CREMA-D}~\cite{cao2014crema} is an audio--visual emotion recognition
dataset comprising 7,442 short video clips from 91 actors across six emotion
categories.
Following the protocol of DGL~\cite{wei2025boosting}, we use 6,698 samples for
training and 744 samples for testing.

\textbf{Kinetics-Sounds (KS)}~\cite{arandjelovic2017look} is an audio--visual
action recognition benchmark derived from Kinetics.
It comprises 34 sound-related action classes and approximately 19,000 video
clips.
We adopt the standardized data splits and training protocol used by
DGL~\cite{wei2025boosting}.

\textbf{URFunny}~\cite{hasan2019ur} is a three-modality humor recognition
benchmark that provides aligned language, acoustic, and visual signals.
We use URFunny to evaluate \regName beyond the two-modality setting and adopt
the preprocessing, backbone, and evaluation protocols used by the corresponding
multimodal baselines.

\textbf{CUB Image-Captions for Clustering (CUBICC)}~\cite{palumbo2024deep}
comprises bird images paired with natural-language descriptions and organizes
the samples into eight semantic categories.
We adopt the implementation and evaluation protocol of
DCMEM~\cite{gaodisentangled} for multimodal clustering.

\textbf{Flickr30K}~\cite{young2014image} comprises 31,783 images, each paired
with five human-written captions.
\textbf{COCO}~\cite{chen2015microsoft} contains a larger collection of complex
everyday scenes, each associated with multiple captions.
We use both datasets to evaluate image--text retrieval with CLIP-based models,
following the training and evaluation protocols described
in~\cite{zhang2024negative,gong2025kernel}.
Retrieval is performed by ranking normalized image and text representations
according to their cross-modal similarity.

\textbf{XRF55}~\cite{wang2024xrf55} is a large-scale multimodal RF sensing
dataset for human activity recognition.
It contains 42,900 synchronized samples from 39 subjects performing 55 action
classes and provides Wi-Fi CSI, RFID, and mmWave radar measurements together
with synchronized visual recordings.
Following the X-Fi framework~\cite{chen2024x}, we assign the first 14 trials of
each action to the training set and the remaining six trials to the test set,
resulting in approximately 30,000 training samples and 12,900 test samples.

\subsection{Training Procedure of \regName}
\label{app:algorithm}

\regName is incorporated into existing multimodal training pipelines as an
auxiliary representation-level regularizer.
It requires no modification to the backbone architecture or inference
procedure.
The task objective is computed using the original modality representations,
whereas the geometric losses operate on their $\ell_2$-normalized versions.
Unless otherwise stated, we use Pareto balancing with $\beta = 0.2$. Fixed coefficients are used only in explicitly labeled ablation and sensitivity experiments (See Fig.~\ref{fig:lambda_sensitivity}).

\paragraph{Baseline implementation and evaluation protocol.}
For each benchmark, all baselines and their \regName-augmented variants use identical data splits, feature extractors, backbone architectures, embedding dimensions, batch sizes, optimizers, and learning-rate schedules; only method-specific components, such as gradient modulation, objective reweighting, or fusion operations, are changed. Whenever available, we use official implementations or publicly released benchmark code; otherwise, we re-implement each method following its original paper and match the reported hyper-parameters as closely as possible. Specifically, CREMA-D, Kinetics-Sounds, and UR-FUNNY follow the DGL protocol~\cite{wei2025boosting}; the X-Fi and COMPASS experiments on XRF55 follow the X-Fi setting~\cite{chen2024x}; CUBICC follows the official DCMEM configuration~\cite{gaodisentangled}; and the CLIP-based retrieval experiments on Flickr30K and COCO follow the NPC pipeline~\cite{zhang2024negative}, with COCO evaluated using the standard 1K test protocol. Across all settings, \regName is applied to the $\ell_2$-normalized outputs of the final modality-specific representation layer, immediately before fusion or the task head; for CLIP, these are the final projected image and text embeddings. Unless otherwise specified, we set $t=1$, use the Pareto-based variant with $\beta=0.2$, and select $\tau$ on the corresponding validation split. All results are obtained using three random seeds (0, 42, and 3407), and we report the mean and standard deviation if possible.

\paragraph{Pareto balancing (Default).}
We additionally consider an adaptive variant that balances the two geometric
terms according to their encoder gradients. Unless otherwise stated, all results denoted by \regName use the
Pareto-balanced variant with $\beta=0.2$. Fixed coefficients $\lambda_d$ and $\lambda_a$ are used only in explicitly labeled ablation and sensitivity experiments.

Let
\begin{equation}
g_{\mathrm d}
=
\nabla_{\theta_e}\mathcal{L}_{\mathrm d},
\qquad
g_{\mathrm a}
=
\nabla_{\theta_e}\mathcal{L}_{\mathrm a},
\end{equation}
where $\theta_e$ denotes the encoder parameters.
We seek the minimum-norm convex combination
$\alpha g_{\mathrm a}+(1-\alpha)g_{\mathrm d}$,
whose closed-form solution is
\begin{equation}
\label{eq:pareto_alpha_closed_form}
\alpha^*
=
\mathrm{clip}_{[0,1]}
\left(
\frac{
\|g_{\mathrm d}\|_2^2
-
\langle g_{\mathrm a},g_{\mathrm d}\rangle
}{
\|g_{\mathrm a}-g_{\mathrm d}\|_2^2
}
\right).
\end{equation}

The resulting geometry gradient is
\begin{equation}
\label{eq:pareto_geom_grad}
g_{\mathrm{geom}}
=
\beta
\left[
\alpha^* g_{\mathrm a}
+
(1-\alpha^*)g_{\mathrm d}
\right],
\end{equation}
where $\beta$ controls the overall regularization strength.
The encoder update then combines this adaptive geometry gradient with the
task gradient:
\begin{equation}
g_{\mathrm{enc}}
=
\nabla_{\theta_e}\mathcal{L}_{\mathrm{task}}
+
g_{\mathrm{geom}}.
\end{equation}

Equivalently, Pareto balancing induces iteration-dependent coefficients
$\lambda_{\mathrm a}^{(t)}=\beta\alpha^*$ and
$\lambda_{\mathrm d}^{(t)}=\beta(1-\alpha^*)$ in
Eq.~\ref{eq:full_objective}.
The predictor parameters remain optimized solely by
$\mathcal{L}_{\mathrm{task}}$, since the geometric regularizers are applied
only to the encoder representations.

\subsection{Training Environment and Computational Overhead}
\label{sec:runtime_overhead}

\noindent
All experiments were conducted on workstations equipped with NVIDIA RTX 4090
GPUs. Unless otherwise specified, each timing experiment was performed under the
same single-GPU setting, with identical software, optimization, and data-loading
configurations for the baseline and its \regName-augmented counterpart.

\regName is implemented as a lightweight training-time regularization module
applied to intermediate modality-wise embeddings. It introduces no additional
trainable parameters and is disabled during inference. The intra-modal
dispersion term involves pairwise distance computations within each modality,
while the inter-modal anchoring term operates on paired cross-modal embeddings.
Both terms are implemented with fully vectorized tensor operations.

\begin{figure}[t]
    \centering
    \includegraphics[width=\linewidth]{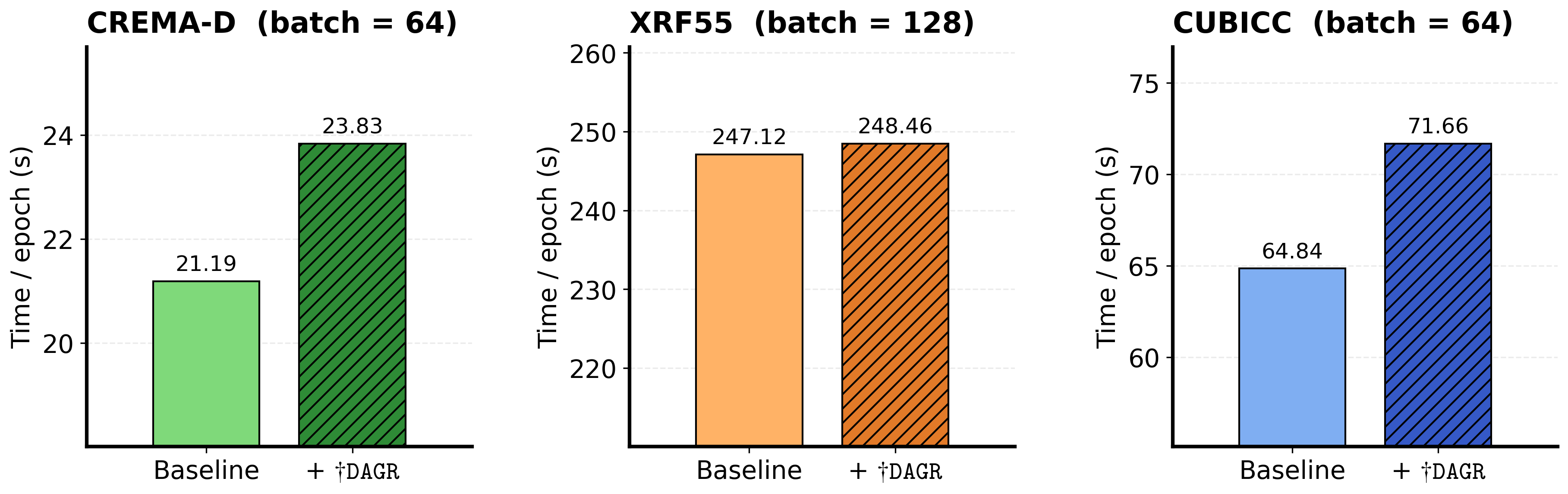}
    \vspace{2mm}
    \caption{
    Runtime comparison between the baseline and \regName across three representative datasets.
    We report CREMA-D and CUBICC with batch size 64, and XRF55 with batch size 128.
    \regName introduces only a small training-time overhead while adding no inference-time cost, since it is applied only during training.
    }
    \label{fig:runtime_overhead}
\end{figure}

Figure~\ref{fig:runtime_overhead} reports the training runtime of the baseline and
\regName under representative feasible batch-size settings on CREMA-D, XRF55, and
CUBICC. The additional cost of \regName is generally modest. In particular, the
runtime on XRF55 remains almost unchanged, while CREMA-D and CUBICC show a moderate
increase. These results indicate that the proposed regularization can be added to
the training objective without substantially changing the computational profile of
the original pipeline. Since \regName is used only during training, it introduces no
additional inference-time cost.

\subsection{Comparison with Representative Multimodal Methods}
\label{app:baseline_scope}

To contextualize the experimental scope of \regName, we summarize
representative multimodal learning methods that are closely related to the
baselines considered in our experiments.
Table~\ref{tab:baseline_scope} reports their evaluated datasets, approximate
dataset scales, reported computing resources, and publication venues.
Because these works address different tasks and use different architectures
and training protocols, the table is intended to provide experimental context
rather than a direct comparison of computational efficiency.

\begin{table*}[t]
\centering
\scriptsize
\setlength{\tabcolsep}{4.2pt}
\renewcommand{\arraystretch}{1.12}

\caption{
Comparison of the experimental scope of representative multimodal learning
methods. Dataset sizes are approximate and computing resources correspond to
those reported in the respective papers when available.
}
\label{tab:baseline_scope}

\begin{tabular}{l|p{5.0cm}|c|c|c}
\toprule
\textbf{Method}
& \textbf{Tested Datasets}
& \textbf{Dataset Scale}
& \textbf{Reported Hardware}
& \textbf{Venue} \\
\midrule

G-Blending~\cite{wang2020makes}
& KS, EPIC-Kitchens, AudioSet
& $\sim$240K--260K
& GPU cluster
& CVPR 2020 \\

OGM-GE~\cite{peng2022balanced}
& CREMA-D, KS, VGGSound
& $\sim$7K--183K
& Single GPU
& CVPR 2022 \\

AGM~\cite{li2023boosting}
& CREMA-D, AV-MNIST, URFunny, MOSEI
& $\sim$7K--16K
& Single GPU
& ICCV 2023 \\

MMPareto~\cite{wei2024mmpareto}
& CREMA-D, KS, CG-MNIST, ModelNet40
& $\sim$7K--19K
& Not reported
& ICML 2024 \\

D\&R~\cite{wei2024diagnosing}
& CREMA-D, KS, UCF-101, MOSI
& $\sim$2K--19K
& 2$\times$ RTX 3090
& ECCV 2024 \\

DCMEM~\cite{gaodisentangled}
& CUBICC, MNIST-SVHN, Spatial Transcriptomics
& $\sim$12K--70K
& RTX 2080 Ti
& NeurIPS 2025 \\

DGL~\cite{wei2025boosting}
& CREMA-D, KS, VGGSound, MOSI, NYUv2
& $\sim$1.4K--183K
& Not reported
& ICCV 2025 \\

X-Fi~\cite{chen2024x}
& XRF55, MM-Fi
& $\sim$43K--320K
& RTX 4090
& ICLR 2025 \\

\midrule
\textbf{\regName (Ours)}
& \textbf{CREMA-D, KS, URFunny, CUBICC, Flickr30K, COCO, XRF55}
& \textbf{$\sim$7K--123K}
& \textbf{RTX 4090}
&  \\

\bottomrule
\end{tabular}
\end{table*}

Compared with prior methods that are typically evaluated within a specific
modality pair or task family, \regName is assessed across audio--visual,
language--audio--visual, image--text, and RF--visual settings.
The experiments additionally cover classification, clustering, and retrieval,
providing evidence that the proposed geometric regularization is not tied to
a particular fusion architecture or modality configuration.

\subsection{Representation-Geometry Diagnostics}
\label{app:geometry_diagnostics}

We define all diagnostics on the fixed evaluation set using the same
representations for every compared method. Let
$\tilde z_i^m\in\mathbb{R}^{d}$ denote the $\ell_2$-normalized representation
of sample $i\in\{1,\ldots,N\}$ from modality $m\in\{1,\ldots,M\}$.
When class labels are available, let $y_i\in\{1,\ldots,C\}$ and define
\begin{equation}
\mathcal I_c=\{i\in\{1,\ldots,N\}:y_i=c\},
\qquad
n_c=|\mathcal I_c|.
\end{equation}
All distances are computed in this normalized representation space.

\paragraph{Between-to-within-class distance ratio.}
For each modality, we compute the class-balanced mean within-class distance
\begin{equation}
D_{\mathrm{within}}^m
=
\frac{1}{C}
\sum_{c=1}^{C}
\frac{1}{\binom{n_c}{2}}
\sum_{\substack{i,j\in\mathcal I_c\\i<j}}
\left\|\tilde z_i^m-\tilde z_j^m\right\|_2,
\label{eq:within_class_distance}
\end{equation}
and the class-balanced mean between-class distance
\begin{equation}
D_{\mathrm{between}}^m
=
\frac{2}{C(C-1)}
\sum_{1\leq c<c'\leq C}
\frac{1}{n_cn_{c'}}
\sum_{i\in\mathcal I_c}
\sum_{j\in\mathcal I_{c'}}
\left\|\tilde z_i^m-\tilde z_j^m\right\|_2.
\label{eq:between_class_distance}
\end{equation}
The reported ratio is
\begin{equation}
\mathrm{B/W}
=
\frac{1}{M}
\sum_{m=1}^{M}
\frac{D_{\mathrm{between}}^m}
{D_{\mathrm{within}}^m+\varepsilon},
\label{eq:bw_ratio}
\end{equation}
where $\varepsilon>0$ is a fixed numerical constant. Larger values indicate
greater inter-class separation relative to intra-class spread.

\paragraph{$k$-nearest-neighbor purity.}
Let $\mathcal N_k^m(i)$ contain the indices of the $k$ nearest neighbors of
$\tilde z_i^m$ among the remaining samples from modality $m$. We define
\begin{equation}
\operatorname{Purity}_k^m
=
\frac{1}{Nk}
\sum_{i=1}^{N}
\sum_{j\in\mathcal N_k^m(i)}
\mathbf{1}[y_j=y_i],
\qquad
\operatorname{Purity}_k
=
\frac{1}{M}
\sum_{m=1}^{M}
\operatorname{Purity}_k^m.
\label{eq:knn_purity}
\end{equation}
The same $k$ is used for all methods within each dataset. Since the
representations are normalized, Euclidean-distance and cosine-similarity
neighbor rankings are equivalent.

\paragraph{Participation-ratio effective rank.}
For each modality, we compute the centered empirical covariance
\begin{equation}
\bar z^m
=
\frac{1}{N}\sum_{i=1}^{N}\tilde z_i^m,
\qquad
\Sigma_m
=
\frac{1}{N-1}
\sum_{i=1}^{N}
(\tilde z_i^m-\bar z^m)
(\tilde z_i^m-\bar z^m)^\top.
\label{eq:empirical_covariance}
\end{equation}
Let $\lambda_{m,1},\ldots,\lambda_{m,d}\geq0$ denote the eigenvalues of
$\Sigma_m$. The modality-specific participation-ratio rank is
\begin{equation}
r_{\mathrm{eff}}(\Sigma_m)
=
\frac{
\left(\sum_{j=1}^{d}\lambda_{m,j}\right)^2
}{
\sum_{j=1}^{d}\lambda_{m,j}^2
},
\label{eq:appendix_effective_rank}
\end{equation}
and we report
\begin{equation}
r_{\mathrm{eff}}
=
\frac{1}{M}
\sum_{m=1}^{M}
r_{\mathrm{eff}}(\Sigma_m).
\label{eq:mean_effective_rank}
\end{equation}
This definition directly matches the spectral quantity analyzed in
Theorem~\ref{thm:dispersion_spectrum}. A larger value indicates a less
concentrated covariance spectrum, but does not by itself imply better
semantic organization.

\paragraph{Paired cross-modal diagnostics.}
Let
\begin{equation}
\mathcal Q
=
\{(m,n):1\leq m<n\leq M\},
\qquad
Q=|\mathcal Q|=\binom{M}{2},
\end{equation}
and define the distance between the paired representations of sample $i$ as
\begin{equation}
d_i^{mn}
=
\left\|\tilde z_i^m-\tilde z_i^n\right\|_2.
\label{eq:paired_modality_distance}
\end{equation}
The mean paired cross-modal drift is
\begin{equation}
D_{\mathrm{drift}}
=
\frac{1}{NQ}
\sum_{i=1}^{N}
\sum_{(m,n)\in\mathcal Q}
d_i^{mn}.
\label{eq:paired_drift}
\end{equation}
This statistic summarizes the average discrepancy between matched
representations. Because \regName enforces bounded rather than maximal
agreement, smaller raw drift is not assumed to be universally preferable.

For a tolerance radius $\tau$, we further report the band-violation rate
\begin{equation}
\rho_\tau
=
\frac{1}{NQ}
\sum_{i=1}^{N}
\sum_{(m,n)\in\mathcal Q}
\mathbf{1}[d_i^{mn}>\tau],
\label{eq:violation_rate}
\end{equation}
and the mean excess violation
\begin{equation}
E_\tau
=
\frac{1}{NQ}
\sum_{i=1}^{N}
\sum_{(m,n)\in\mathcal Q}
(d_i^{mn}-\tau)_+^2,
\label{eq:excess_violation}
\end{equation}
where $(u)_+=\max(u,0)$. The former measures how frequently matched pairs
fall outside the agreement band, whereas the latter additionally measures
the magnitude of those violations. For baselines without a training-time
tolerance, both quantities are evaluated using the same $\tau$ as the
corresponding \regName configuration. For $M>2$, all three cross-modal
statistics average over every unordered modality pair in $\mathcal Q$.

These diagnostics capture complementary properties. B/W and kNN purity
measure label-consistent semantic organization, $r_{\mathrm{eff}}$ measures
spectral spread, and $D_{\mathrm{drift}}$, $\rho_\tau$, and $E_\tau$
characterize paired cross-modal discrepancy. Since $E_\tau$ directly mirrors
the agreement penalty optimized by \regName, we treat it as a mechanism
diagnostic rather than independent evidence of semantic quality. In
particular, the label-agnostic dispersion term alone is not assumed to ensure
class compactness; all geometry diagnostics are therefore interpreted jointly
with downstream task performance.

\section{Additional Experiments and More Visualizations}
\label{app:experiments}

\subsection{Hyper-parameter Sensitivity}
\label{app:hp}

Our method introduces two weighting coefficients, $\lambda_{\mathbf{d}}$ and $\lambda_{\mathbf{a}}$, to balance intra-modal compactness and inter-modal alignment. In principle, tuning these two hyper-parameters also depends on the hinge loss threshold $\tau$, since the effective contribution of the hinge term is jointly determined by both the margin $\tau$ and the loss weights. As a result, naively performing a grid search over $(\lambda_{\mathbf{d}}, \lambda_{\mathbf{a}}, \tau)$ would lead to a large and unstable hyper-parameter space.

\begin{figure}[t]
    \centering
    \begin{minipage}{0.48\linewidth}
        \centering
        \includegraphics[width=\linewidth]{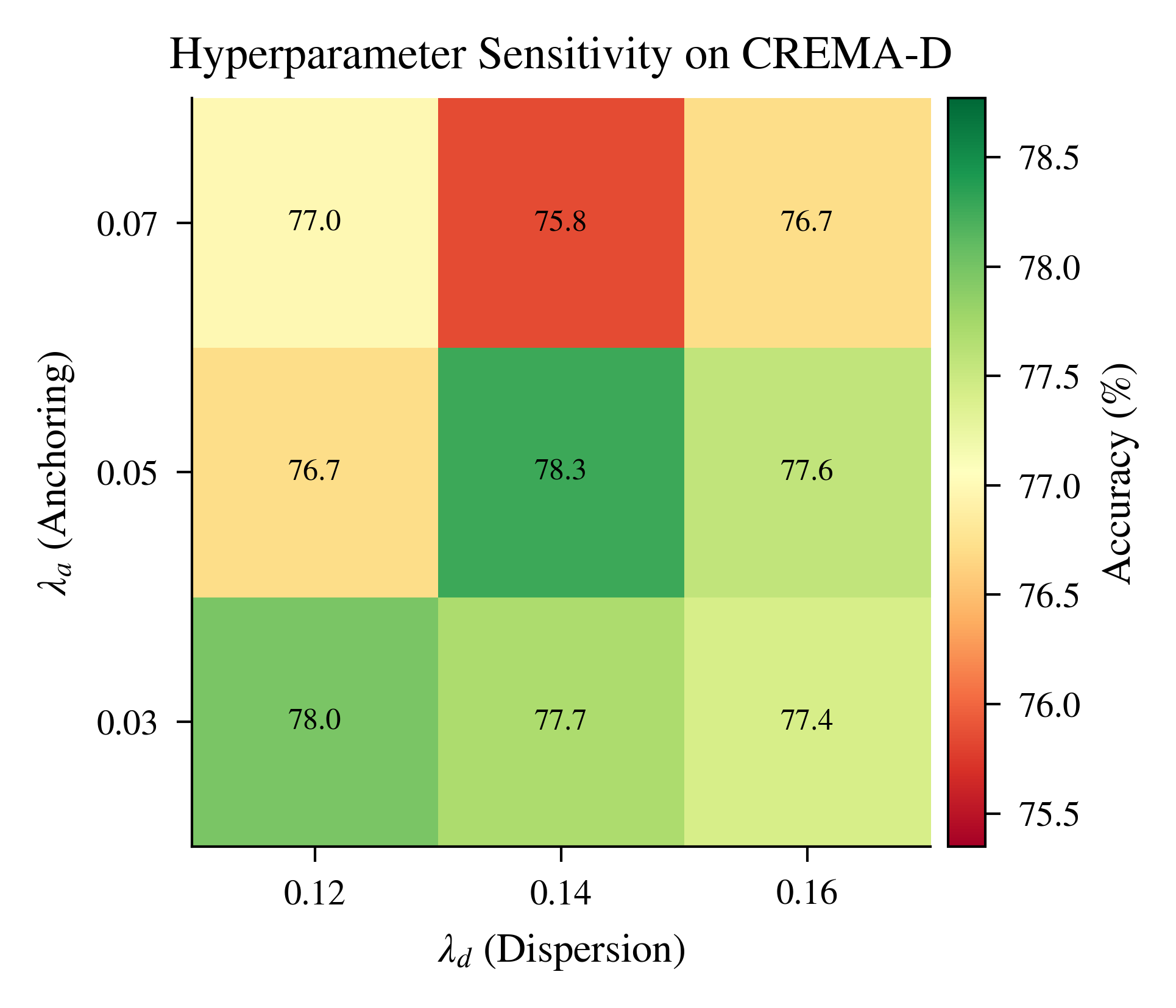}
    \end{minipage}
    \hfill
    \begin{minipage}{0.48\linewidth}
        \centering
        \includegraphics[width=\linewidth]{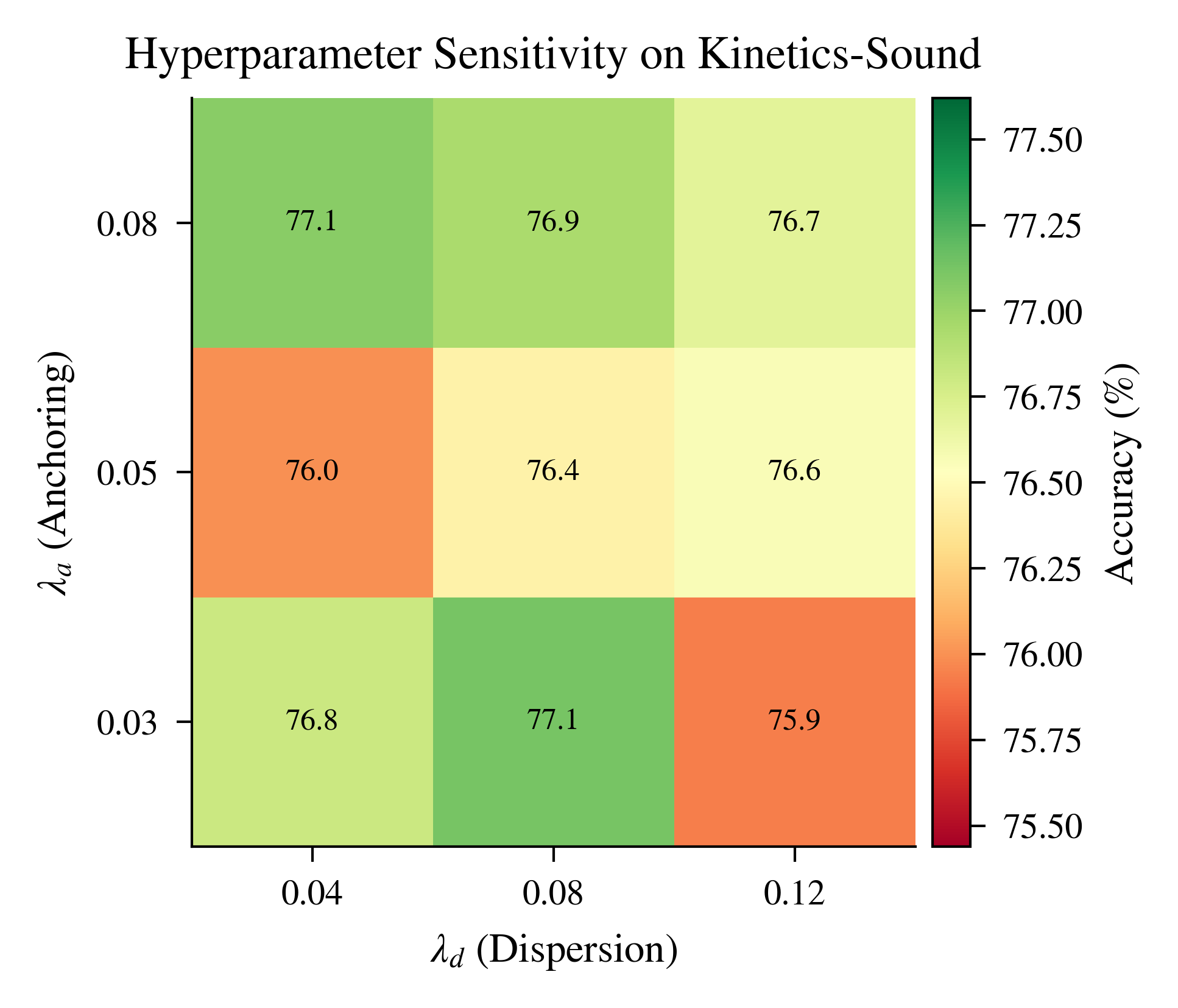}
    \end{minipage}
    \caption{Sensitivity analysis of $\lambda_{\mathbf{d}}$ and $\lambda_{\mathbf{a}}$ with the hinge threshold fixed to $\tau=0$. Left: CREMA-D. Right: Kinetics-Sound.}
    \label{fig:lambda_sensitivity}
\end{figure}

\paragraph{Sensitivity to $\lambda_{\mathbf{d}}$ and $\lambda_{\mathbf{a}}$.}
To first isolate the effect of the two weighting coefficients, we fix the hinge threshold to $\tau=0$ and conduct a two-dimensional grid search over $(\lambda_{\mathbf{d}}, \lambda_{\mathbf{a}})$. Figure~\ref{fig:lambda_sensitivity} shows the resulting performance heatmaps on CREMA-D and Kinetics-Sound.

From the results, we observe that moderate values of both $\lambda_{\mathbf{d}}$ and $\lambda_{\mathbf{a}}$ generally lead to better performance, while overly large weights do not consistently improve accuracy. This suggests that excessively emphasizing the auxiliary contrastive objectives may dominate the optimization process and suppress task-relevant discriminative cues learned by the primary classification loss. Moreover, strong intra-/inter-regularization may over-constrain the feature geometry, reducing representation flexibility and harming generalization.

\paragraph{Joint sensitivity of $\beta$ and $\tau$.}
Motivated by the above observations, we further adopt a Pareto-style parameterization that controls the total regularization strength using a single scalar $\beta = \lambda_{\mathbf{d}} + \lambda_{\mathbf{a}}$, while maintaining a balanced trade-off between intra-modal compactness and inter-modal alignment. Specifically, we consider $\beta \in \{0.1, 0.15, 0.2\}$, which significantly reduces the hyper-parameter search space and improves training stability.

\begin{figure}[t]
    \centering
    \begin{minipage}{0.48\linewidth}
        \centering
        \includegraphics[width=\linewidth]{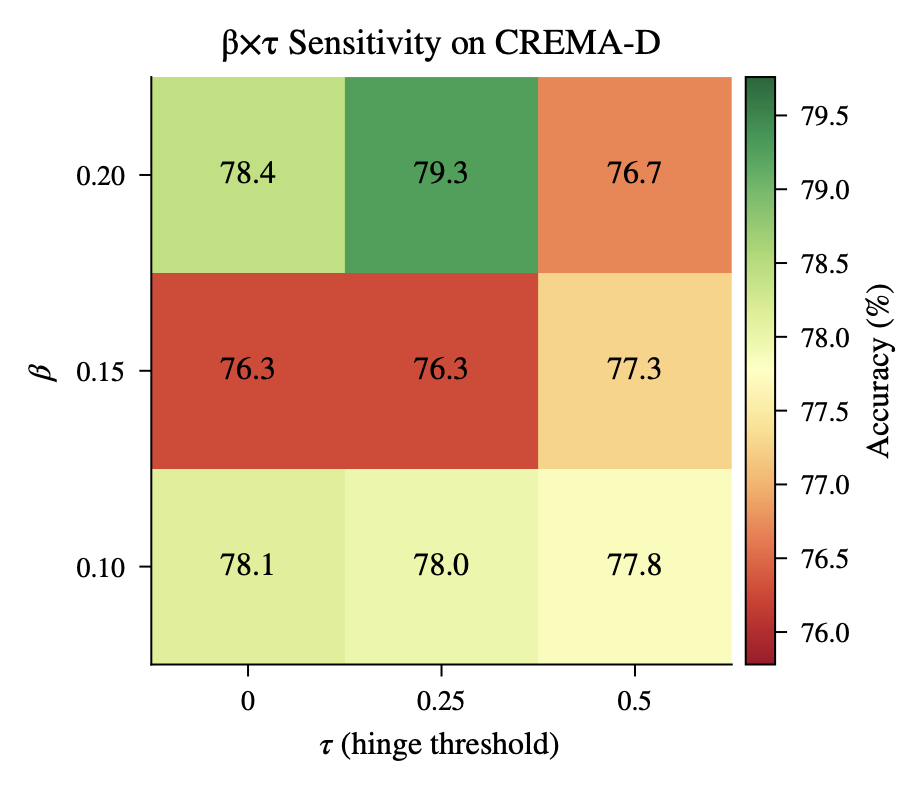}
    \end{minipage}
    \hfill
    \begin{minipage}{0.48\linewidth}
        \centering
        \includegraphics[width=\linewidth]{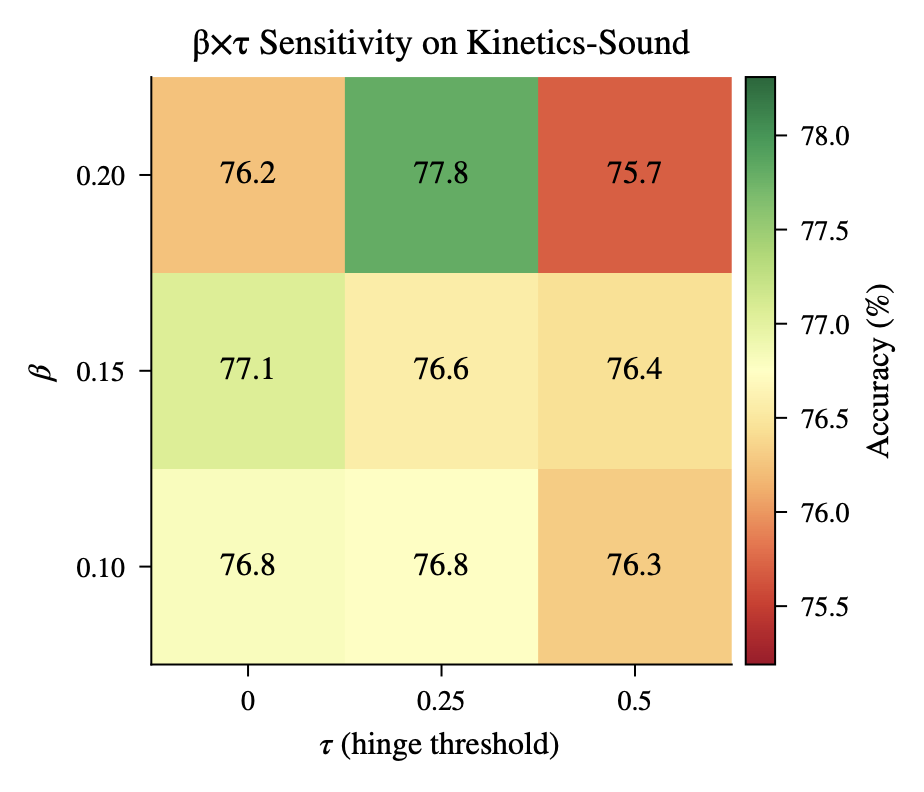}
    \end{minipage}
    \caption{Joint sensitivity analysis of total regularization strength $\beta$ and hinge threshold $\tau$ across two datasets. Left: CREMA-D. Right: Kinetics-Sound.}
    \label{fig:beta_tau_sensitivity}
\end{figure}

We then conduct an ablation study on the hinge threshold $\tau \in \{0, 0.25, 0.5\}$ under different values of $\beta$. Figure~\ref{fig:beta_tau_sensitivity} visualizes the joint sensitivity of $\beta$ and $\tau$ on CREMA-D and Kinetics-Sound. Overall, the results indicate that our method is relatively robust to moderate variations of $\tau$, while excessively large margins may degrade performance when combined with strong regularization. This further supports our choice of using a small $\beta$ together with a mild hinge threshold.

\subsection{Robustness to Noisy Image--Text Correspondences}
\label{sec:robustness:noisy_image_text}

We further evaluate the robustness of \regName under corrupted cross-modal supervision, following the noisy image--text correspondence setting of~\cite{zhang2024negative}. During training, a controlled fraction of image--text pairs is randomly mismatched, while retrieval is evaluated on the clean Flickr30K and COCO test splits. This setting complements the modality-imbalance experiment above by corrupting
image--text correspondences rather than removing one modality. We compare CLIP and CLIP+\regName{} under
different noise ratios and report Recall@\{1,5,10\} for both image-to-text and text-to-image retrieval.

As shown in Table~\ref{tab:clip-noisy-retrieval}, CLIP+\regName{} consistently improves over vanilla CLIP
across both datasets, all noise ratios, and both retrieval directions. The improvement becomes particularly
pronounced under severe correspondence noise. At 60\% noise, CLIP+\regName{} improves I2T/T2I R@1 by
15.1/14.6 points on Flickr30K and by 9.1/12.1 points on COCO. These results suggest that geometry-aware
regularization helps stabilize multimodal retrieval representations when image--text correspondences are
substantially corrupted.

\begin{table*}[t]
\centering
\caption{
\textbf{Robustness to noisy image--text correspondences on Flickr30K~\cite{young2014image} and COCO~\cite{chen2015microsoft}.}
We compare CLIP and CLIP+\regName{} under different correspondence noise ratios.
Numbers are R@\{1,5,10\} percentages.
}
\label{tab:clip-noisy-retrieval}
\vspace{0.35em}

\footnotesize
\setlength{\tabcolsep}{4.8pt}
\renewcommand{\arraystretch}{1.08}

\textbf{Image-to-Text Retrieval}
\vspace{0.25em}

\begin{tabular*}{\textwidth}{@{\extracolsep{\fill}}llcccccc@{}}
\toprule
\multirow{2}{*}{Noise Ratio}
& \multirow{2}{*}{Method}
& \multicolumn{3}{c}{Flickr30K}
& \multicolumn{3}{c}{COCO} \\
\cmidrule(lr){3-5}\cmidrule(lr){6-8}
& & R@1 & R@5 & R@10 & R@1 & R@5 & R@10 \\
\midrule
\multirow{2}{*}{0\%}
& CLIP
& \second{86.2} & \second{97.6} & \second{99.2}
& \second{79.9} & \second{95.1} & \second{98.1} \\
& CLIP+\regName{}
& \best{87.2}
& \best{98.1}
& \best{99.5}
& \best{81.9}
& \best{96.2}
& \best{98.8} \\
\midrule
\multirow{2}{*}{20\%}
& CLIP
& \second{82.3} & \second{95.5} & \second{98.3}
& \second{75.0} & \second{93.1} & \second{97.2} \\
& CLIP+\regName{}
& \best{85.3}
& \best{97.0}
& \best{99.2}
& \best{78.4}
& \best{95.5}
& \best{98.2} \\
\midrule
\multirow{2}{*}{40\%}
& CLIP
& \second{76.2} & \second{93.3} & \second{96.5}
& \second{70.7} & \second{91.7} & \second{96.2} \\
& CLIP+\regName{}
& \best{84.4}
& \best{96.2}
& \best{98.6}
& \best{78.1}
& \best{95.2}
& \best{98.0} \\
\midrule
\multirow{2}{*}{60\%}
& CLIP
& \second{66.3} & \second{87.3} & \second{93.0}
& \second{67.0} & \second{88.8} & \second{95.0} \\
& CLIP+\regName{}
& \best{81.4}
& \best{95.7}
& \best{98.3}
& \best{76.1}
& \best{94.3}
& \best{97.6} \\
\bottomrule
\end{tabular*}

\vspace{0.8em}
\textbf{Text-to-Image Retrieval}
\vspace{0.25em}

\begin{tabular*}{\textwidth}{@{\extracolsep{\fill}}llcccccc@{}}
\toprule
\multirow{2}{*}{Noise Ratio}
& \multirow{2}{*}{Method}
& \multicolumn{3}{c}{Flickr30K}
& \multicolumn{3}{c}{COCO} \\
\cmidrule(lr){3-5}\cmidrule(lr){6-8}
& & R@1 & R@5 & R@10 & R@1 & R@5 & R@10 \\
\midrule
\multirow{2}{*}{0\%}
& CLIP
& \second{72.9} & \second{92.3} & \second{96.0}
& \second{65.0} & \second{90.3} & \second{98.1} \\
& CLIP+\regName{}
& \best{75.1}
& \best{93.6}
& \best{96.9}
& \best{68.0}
& \best{92.1}
& \best{98.8} \\
\midrule
\multirow{2}{*}{20\%}
& CLIP
& \second{66.0} & \second{88.5} & \second{93.5}
& \second{58.7} & \second{86.1} & \second{97.2} \\
& CLIP+\regName{}
& \best{71.6}
& \best{91.8}
& \best{96.0}
& \best{65.5}
& \best{90.5}
& \best{98.2} \\
\midrule
\multirow{2}{*}{40\%}
& CLIP
& \second{59.4} & \second{85.0} & \second{90.9}
& \second{54.7} & \second{83.4} & \second{96.2} \\
& CLIP+\regName{}
& \best{71.0}
& \best{90.7}
& \best{95.0}
& \best{64.0}
& \best{89.8}
& \best{98.0} \\
\midrule
\multirow{2}{*}{60\%}
& CLIP
& \second{52.1} & \second{78.8} & \second{87.4}
& \second{49.7} & \second{79.6} & \second{95.0} \\
& CLIP+\regName{}
& \best{66.7}
& \best{88.9}
& \best{93.7}
& \best{61.8}
& \best{88.1}
& \best{97.6} \\
\bottomrule
\end{tabular*}

\vspace{-1mm}
\end{table*}

\label{sec:robustness}

\subsection{Robustness to Missing or Corrupted Modalities}
\label{sec:robustness}

We further evaluate \regName under \emph{test-time modality degradation} to directly validate the motivation in Sec.~\ref{sec:introduction}:
when one modality becomes partially missing or corrupted, a robust multimodal model should degrade \emph{gracefully}.
Unless otherwise specified, we corrupt \emph{only one} modality at a time during inference while keeping all other modalities unchanged.

\paragraph{CREMA-D: missing and corruption stress tests.}
For the audio-visual setting on CREMA-D, we consider both \emph{missingness} and \emph{sensor-like corruptions}.
(i) \textbf{Random missingness.} With probability $p\in[0,1]$, we replace a modality with a missing placeholder at test time (feature-level masking),
resulting in missing-audio and missing-visual curves.
(ii) \textbf{Audio noise.} We inject additive noise with different SNR levels (dB) to simulate degraded audio acquisition.
(iii) \textbf{SpecAugment / frame-drop / cutout.} We apply standard modality-specific corruptions:
SpecAugment masks a fraction of time/frequency bins on audio features,
frame-drop randomly removes a fraction of frames from the visual stream,
and cutout occludes a random rectangular region with a given area ratio.

\begin{figure}[t]
    \centering
    \includegraphics[width=0.98\linewidth]{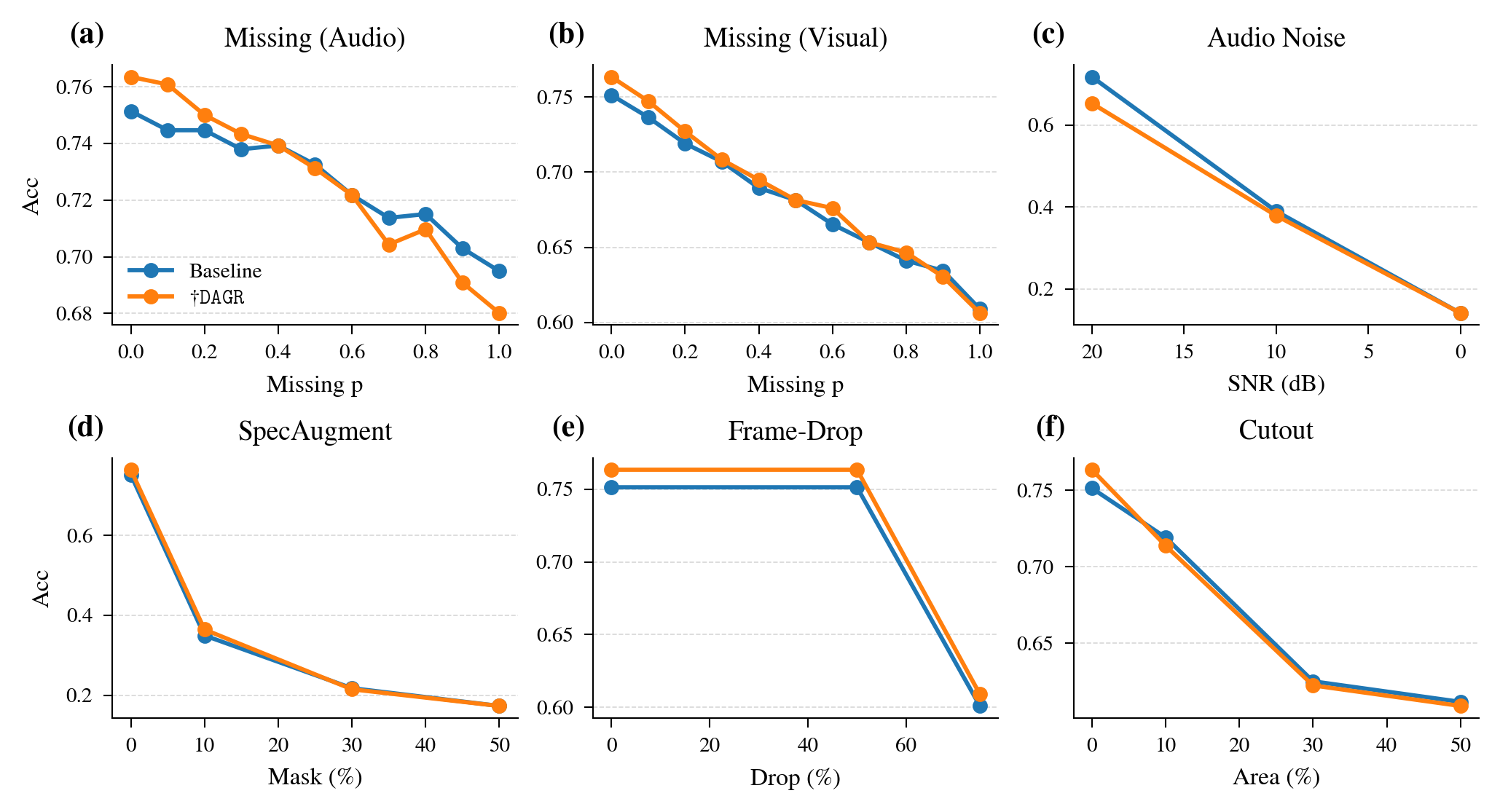}
    \caption{\textbf{Robustness under missing or corrupted modalities on CREMA-D.}
    We evaluate test-time degradation by (a) missing audio, (b) missing visual, (c) additive audio noise (SNR sweep),
    and modality-specific corruptions including (d) SpecAugment, (e) frame-drop, and (f) cutout.
    \regName generally exhibits improved robustness in the low-to-moderate degradation regime and maintains competitive performance under severe corruption,
    yielding smoother degradation trends than the baseline.}
    \label{fig:cremad_robustness}
\end{figure}

\paragraph{X-FI: dropout corruption.}
We first evaluate robustness to \emph{partial feature loss} by progressively applying random dropout to a single modality at test time,
while keeping all other modalities unchanged.
Specifically, for the selected modality we randomly mask a fraction $\rho$ of its feature dimensions or tokens (drop ratio $\rho\in[0,1]$) before inference.
This setting simulates scenarios where a modality is only partially available due to occlusion, packet loss, or unstable sensing.

\begin{figure}[t]
    \centering
    \includegraphics[width=0.32\linewidth]{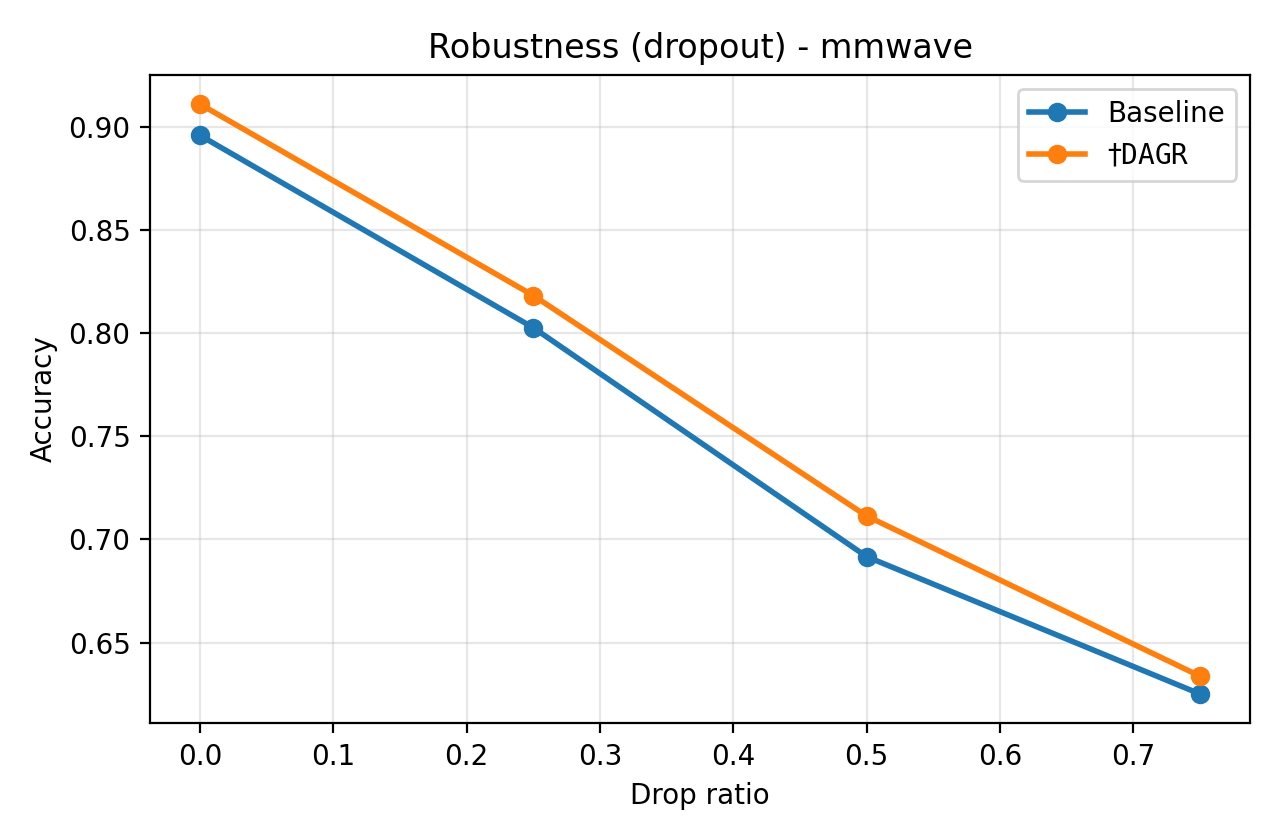}\hfill
    \includegraphics[width=0.32\linewidth]{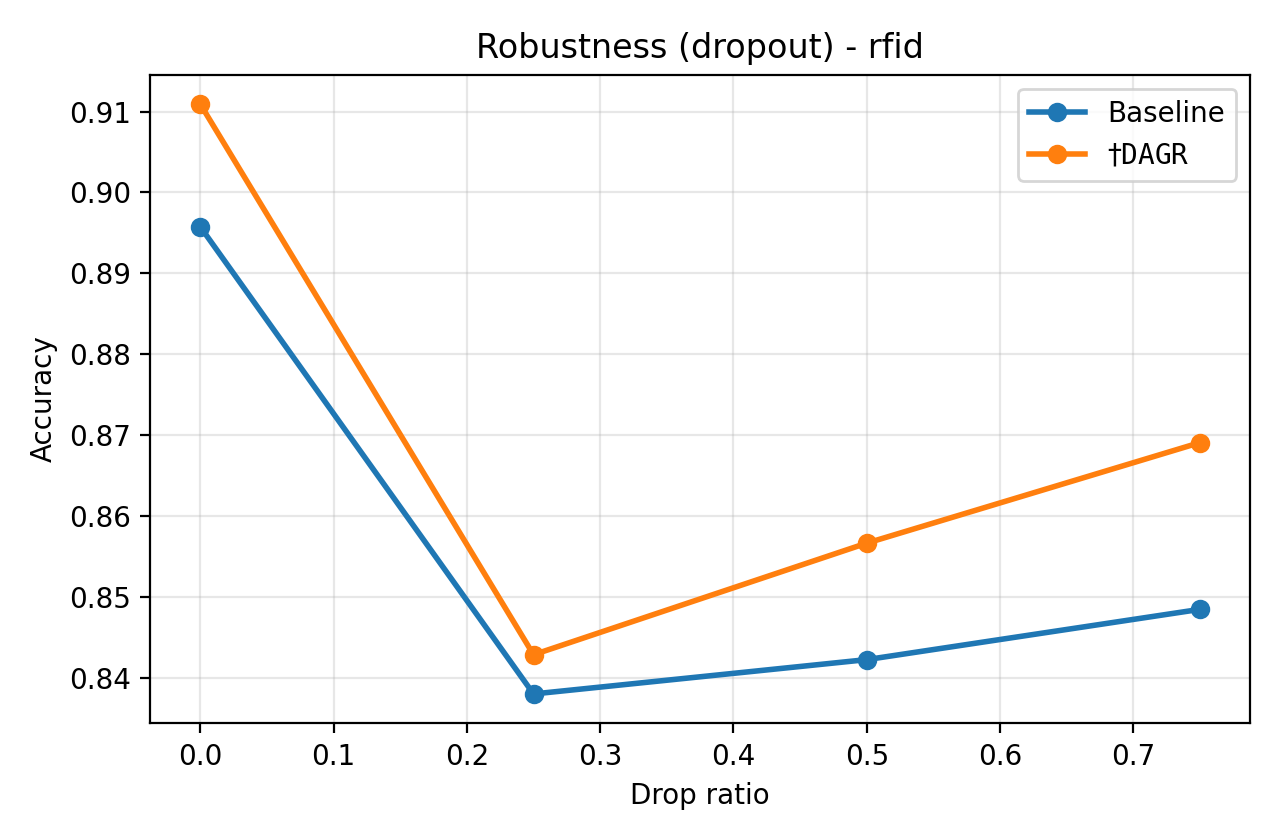}\hfill
    \includegraphics[width=0.32\linewidth]{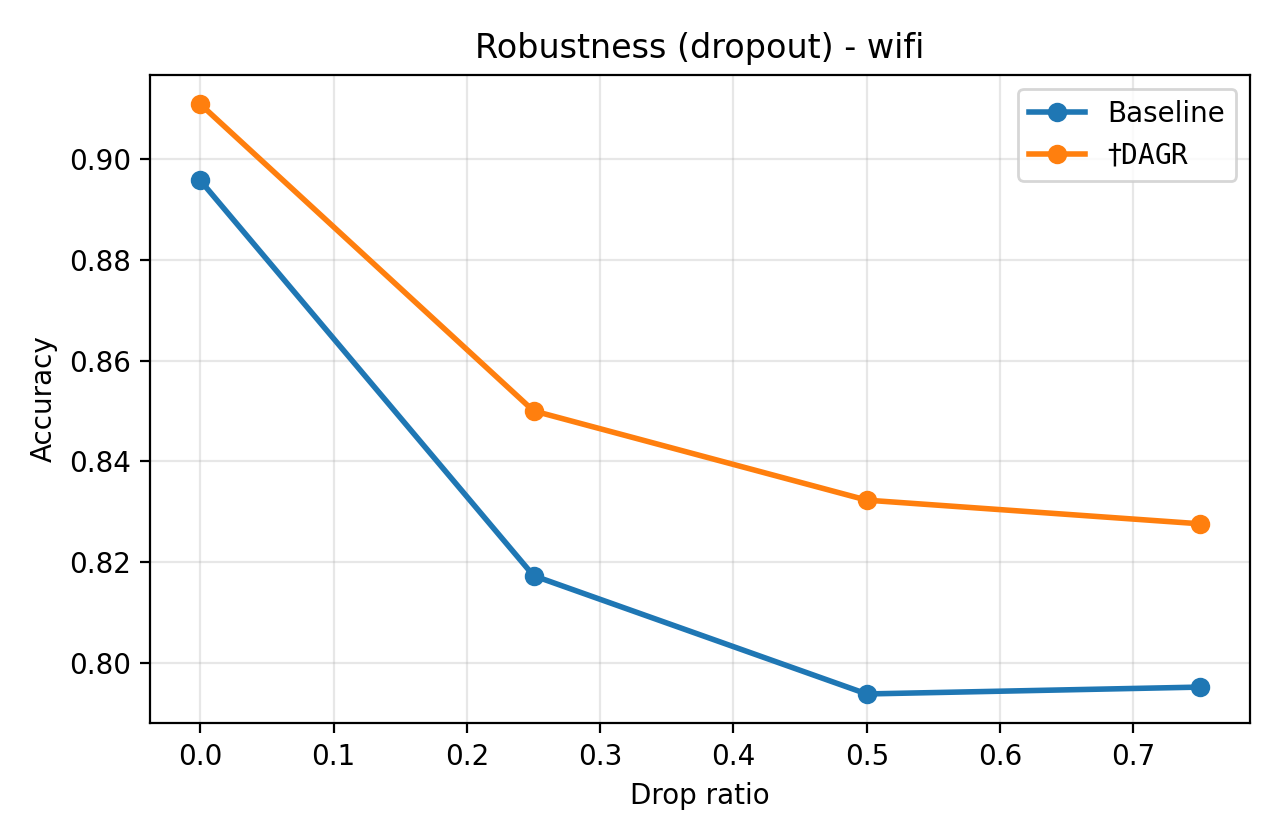}
    \vspace{2mm}
    \caption{\textbf{Robustness under dropout on X-FI.}
    We progressively drop a fraction $\rho$ of features from one modality at test time, while keeping the other modalities intact.
    From left to right: mmWave, RFID, and WiFi.
    \regName attains higher accuracy and degrades more gracefully than the baseline, especially under moderate-to-severe dropout.}
    \label{fig:xfi_dropout}
\end{figure}

\paragraph{X-FI: Gaussian corruption.}
Next, we study robustness to \emph{additive sensor noise} by injecting Gaussian noise into one modality at test time.
For modality $m$, the corrupted input is constructed as
$\tilde{\mathbf{x}}^{(m)}=\mathbf{x}^{(m)}+\sigma\cdot s^{(m)}\epsilon$,
where $\epsilon\sim\mathcal{N}(0,1)$, $\sigma$ controls the noise strength, and $s^{(m)}$ is a modality-specific scale to normalize noise magnitude across modalities.
We sweep $\sigma$ from weak to strong corruption and report the resulting degradation curves.

\begin{figure}[t]
    \centering
    \includegraphics[width=0.32\linewidth]{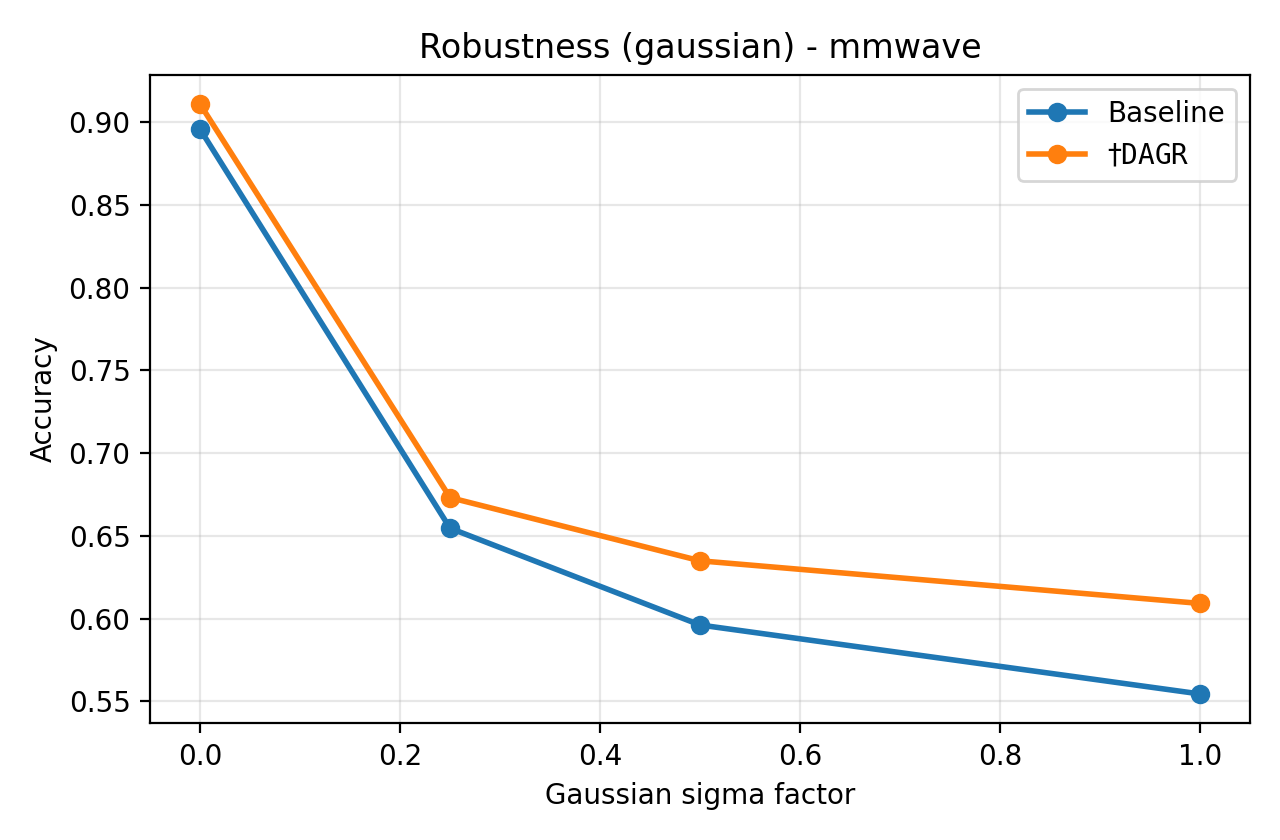}\hfill
    \includegraphics[width=0.32\linewidth]{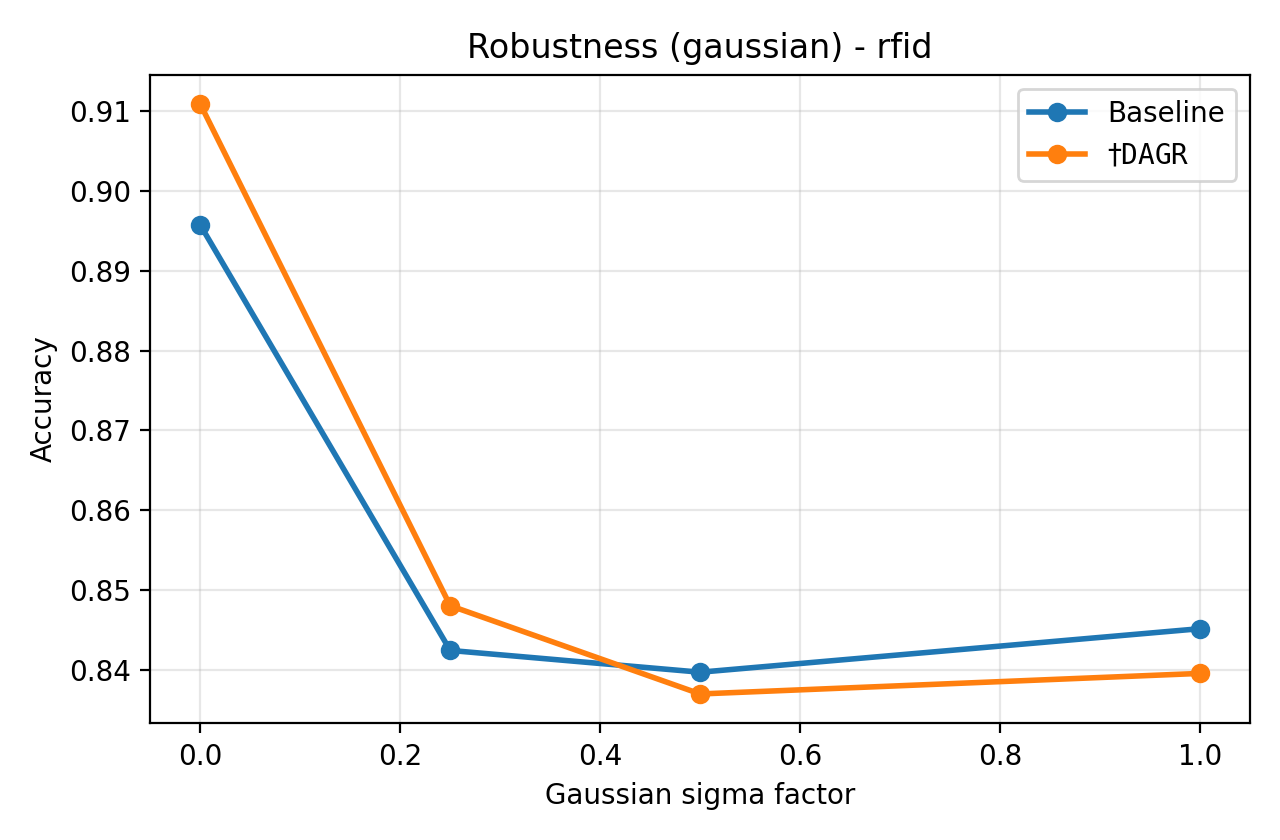}\hfill
    \includegraphics[width=0.32\linewidth]{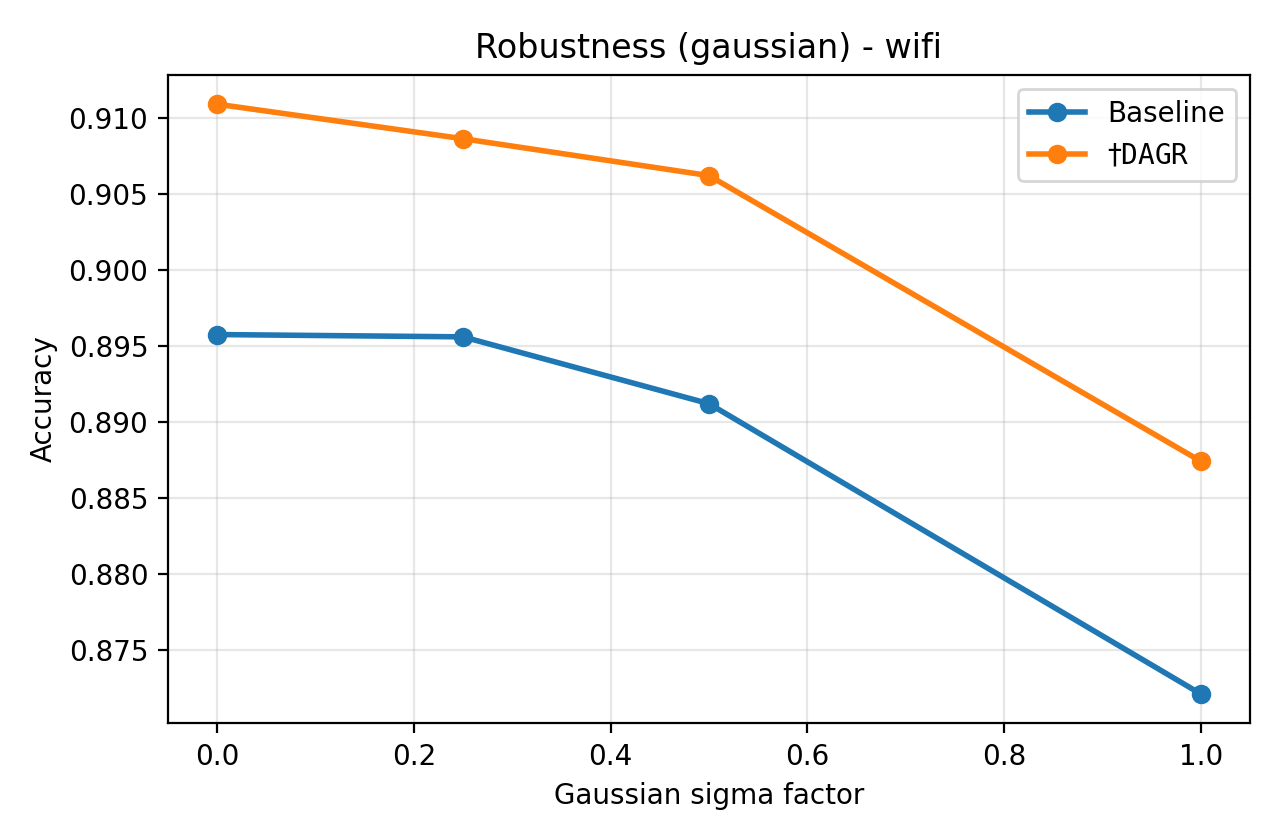}
    \vspace{2mm}
    \caption{\textbf{Robustness under Gaussian noise on X-FI.}
    We inject additive Gaussian noise with factor $\sigma$ into a single modality at test time.
    From left to right: mmWave, RFID, and WiFi.
    \regName shows improved robustness under noisy inputs (notably for mmWave and WiFi) and generally degrades more smoothly than the baseline as noise increases.}
    \label{fig:xfi_gaussian}
\end{figure}

\paragraph{X-FI: random missingness.}
Finally, we evaluate robustness to \emph{temporal or segment-level modality absence} by randomly masking parts of a modality.
Concretely, for the selected modality, each temporal step or segment is independently removed with probability $p$ (drop rate $p\in[0,1]$),
while all other modalities remain intact.
This setting simulates intermittent sensor failures or missing data streams, which are common in real-world deployments.

\begin{figure}[t]
    \centering
    \includegraphics[width=0.32\linewidth]{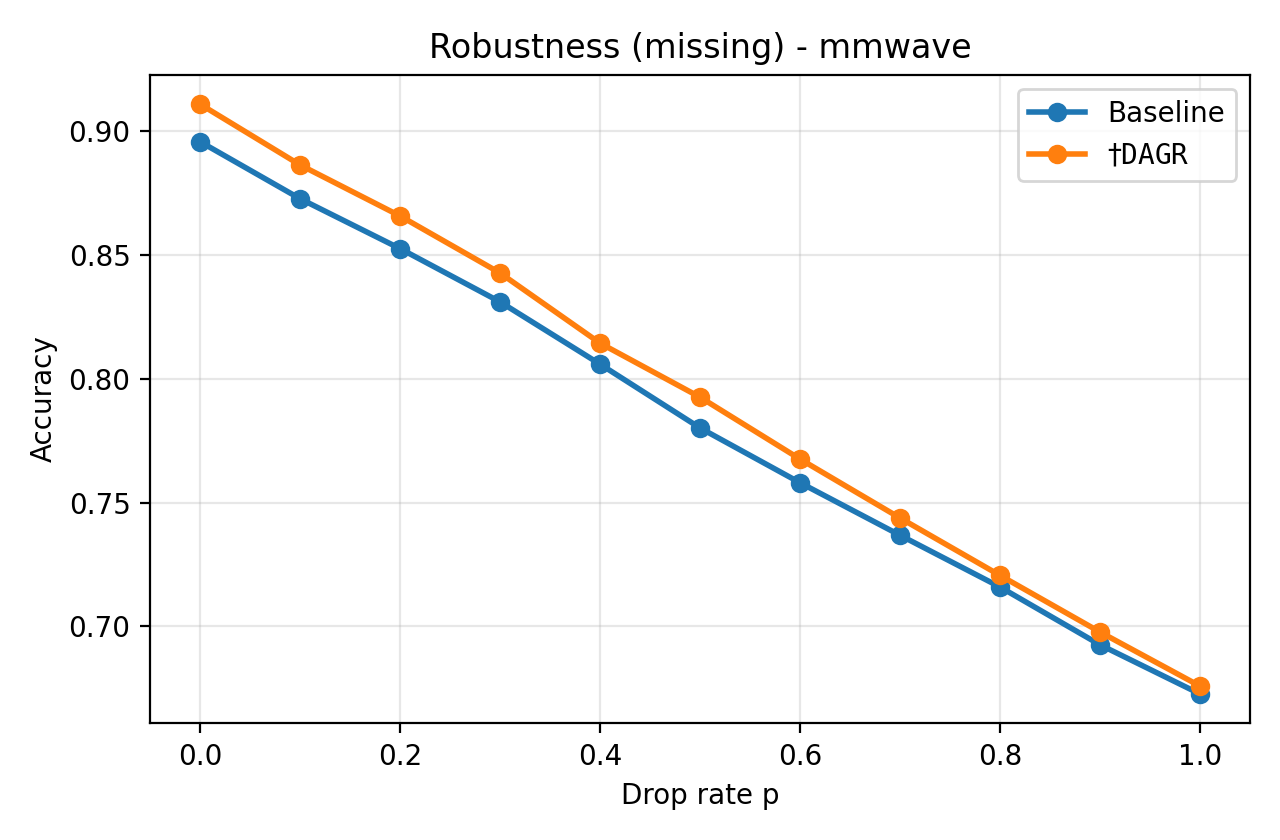}\hfill
    \includegraphics[width=0.32\linewidth]{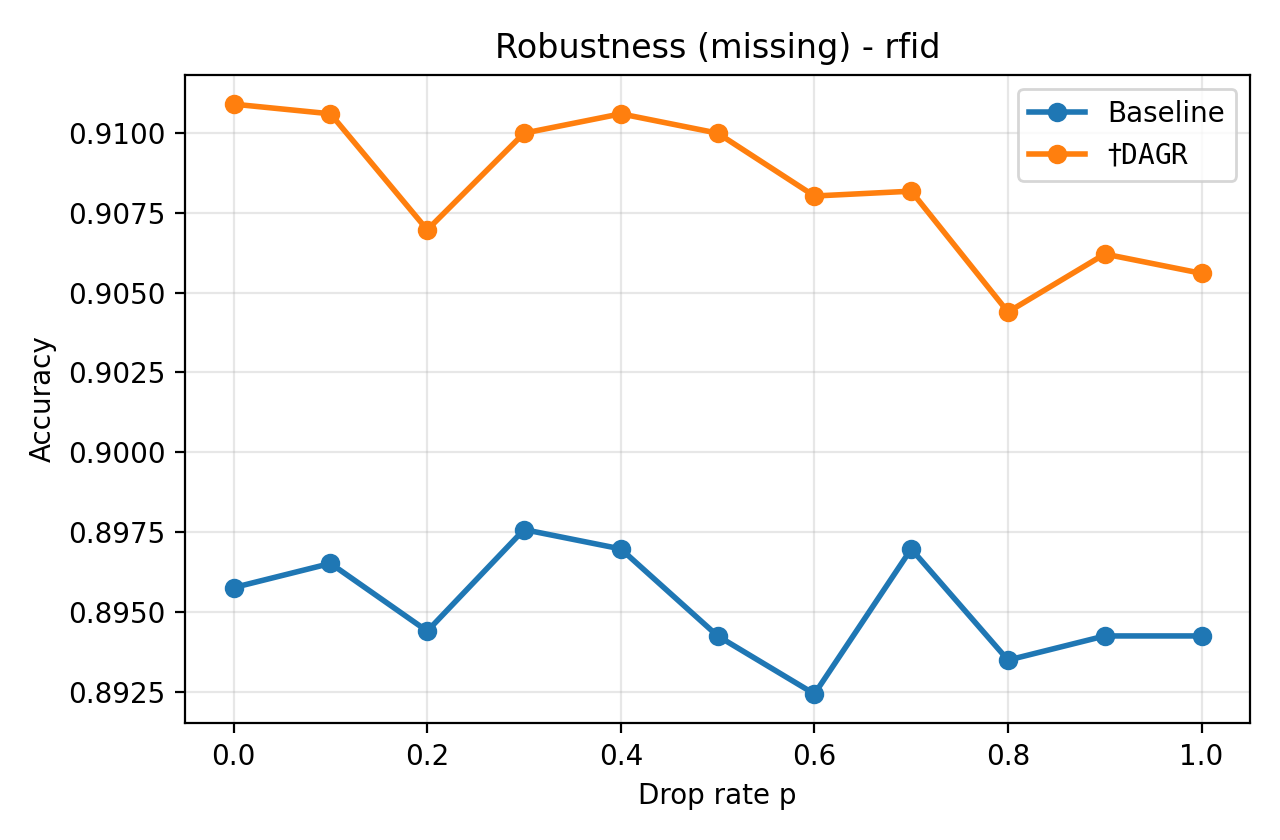}\hfill
    \includegraphics[width=0.32\linewidth]{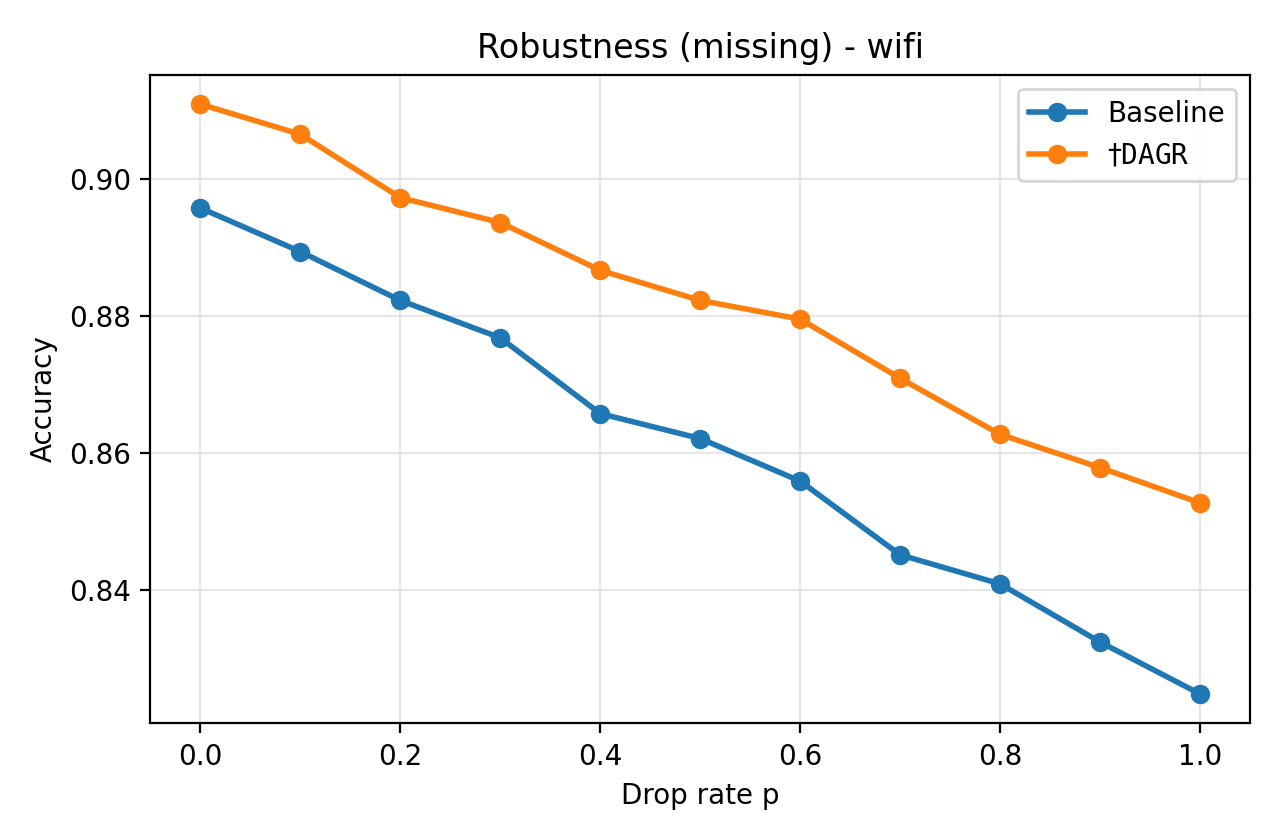}
    \vspace{2mm}
    \caption{\textbf{Robustness under random missingness on X-FI.}
    We randomly mask temporal steps or segments of one modality with probability $p$ at test time.
    From left to right: mmWave, RFID, and WiFi.
    \regName maintains higher accuracy and better stability under increasing modality missingness.}
    \label{fig:xfi_missing}
\end{figure}

\paragraph{Discussion.}
Fig.~\ref{fig:cremad_robustness} and Figs.~\ref{fig:xfi_dropout}--\ref{fig:xfi_missing} provide a direct robustness diagnosis beyond standard unimodal ablations.
Across datasets, performance degrades monotonically as corruption severity increases, validating these perturbations as meaningful stress tests.
Overall, \regName tends to achieve higher accuracy and smoother degradation trends when a single modality becomes unreliable, indicating improved robustness of multimodal fusion.

On X-FI, the gains are particularly pronounced for mmWave and WiFi under moderate-to-severe corruption, where the baseline suffers sharper drops.
For RFID, partially corrupted inputs can be more harmful than fully missing ones, suggesting that misleading low-quality signals may interfere with fusion;
in these regimes, \regName maintains a clearer advantage, consistent with stronger resistance to unreliable modality information.
On CREMA-D, \regName improves robustness in the low-to-moderate degradation regime across multiple corruption types and remains competitive under severe corruption,
supporting the claim that \regName alleviates geometric pathologies that otherwise amplify sensitivity to modality degradation.

\subsection{Additional Qualitative Analyses}
\label{app:additional_qualitative}

We provide additional qualitative visualizations on Kinetics-Sounds, CUBICC, and CLIP-based image--text retrieval. These results illustrate how the learned representations and attention patterns change after
introducing \regName. Because low-dimensional projections and selected attention maps can depend on visualization choices, we treat them as supplementary illustrations rather than standalone evidence of improved
representation quality.

\begin{figure*}[t]
    \centering
    \includegraphics[width=0.98\textwidth]{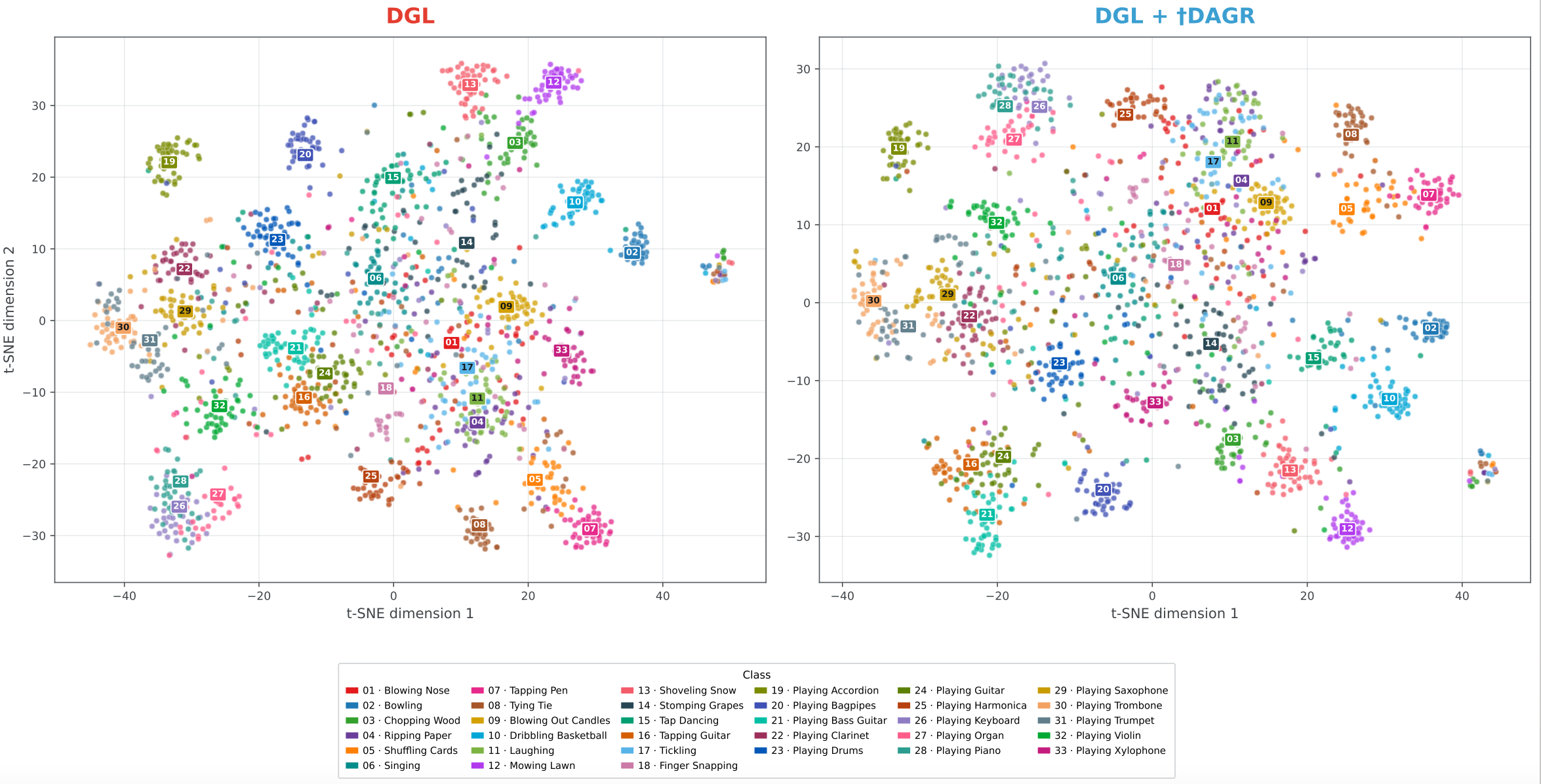}
    \caption{
        \textbf{Class-wise t-SNE visualization on Kinetics-Sounds.}
        We compare representations learned by DGL (left) and
        DGL+\regName{} (right), using the same evaluation samples and
        visualization settings. Colors and numerical labels denote action
        classes. The visualization is intended as a qualitative complement
        to the quantitative results.
    }
    \label{fig:tsne_ks}
\end{figure*}

\paragraph{Class-wise geometry on Kinetics-Sounds.}
Figure~\ref{fig:tsne_ks} presents t-SNE visualizations of the representations learned by DGL and DGL+\regName using the same evaluation samples and visualization settings. Both methods form several class-specific clusters while retaining a substantially mixed central region. Compared with DGL, DGL+\regName preserves compact peripheral clusters and exhibits modestly clearer local grouping for several categories, although considerable overlap remains among other classes. We therefore interpret this visualization as supplementary qualitative evidence rather than a standalone measure of representation quality.

\begin{figure}[t]
    \centering
    \includegraphics[width=0.95\linewidth]{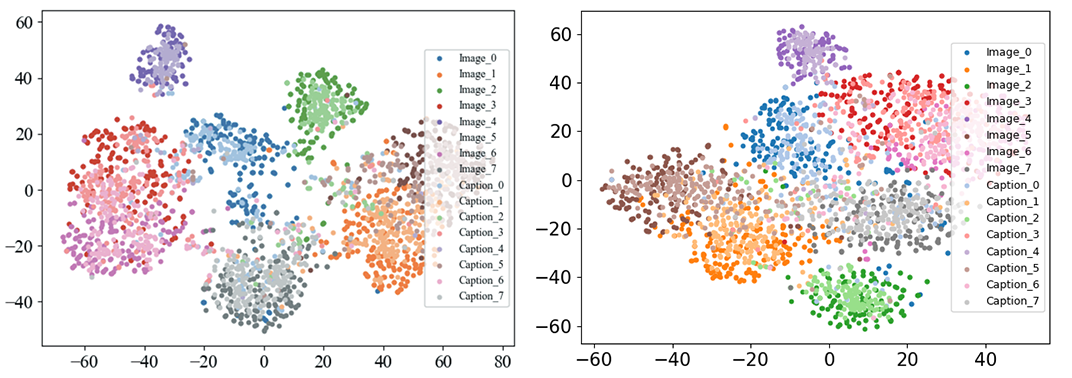}
    \vspace{2mm}
    \caption{
    t-SNE visualization of multimodal embeddings on CUBICC.
    \regName improves semantic compactness and stabilizes image--caption alignment relative to the baseline.
    }
    \label{fig:tsne_cubicc}
\end{figure}

\paragraph{Joint image--text geometry on CUBICC.}
Figure~\ref{fig:tsne_cubicc} presents the visualization results on the CUBICC dataset.
Although the baseline already shows partial semantic separation, image and caption embeddings from the same
category often form loose, fragmented, or only weakly overlapping structures. With \regName, the clusters become
more coherent and better separated across categories. Moreover, paired image--caption embeddings exhibit stronger
within-category overlap, indicating improved cross-modal consistency. These results suggest that \regName enhances
the semantic organization of multimodal representations without imposing a rigid global alignment between modalities.

\begin{figure*}[t]
    \centering
    \includegraphics[width=\textwidth]
    {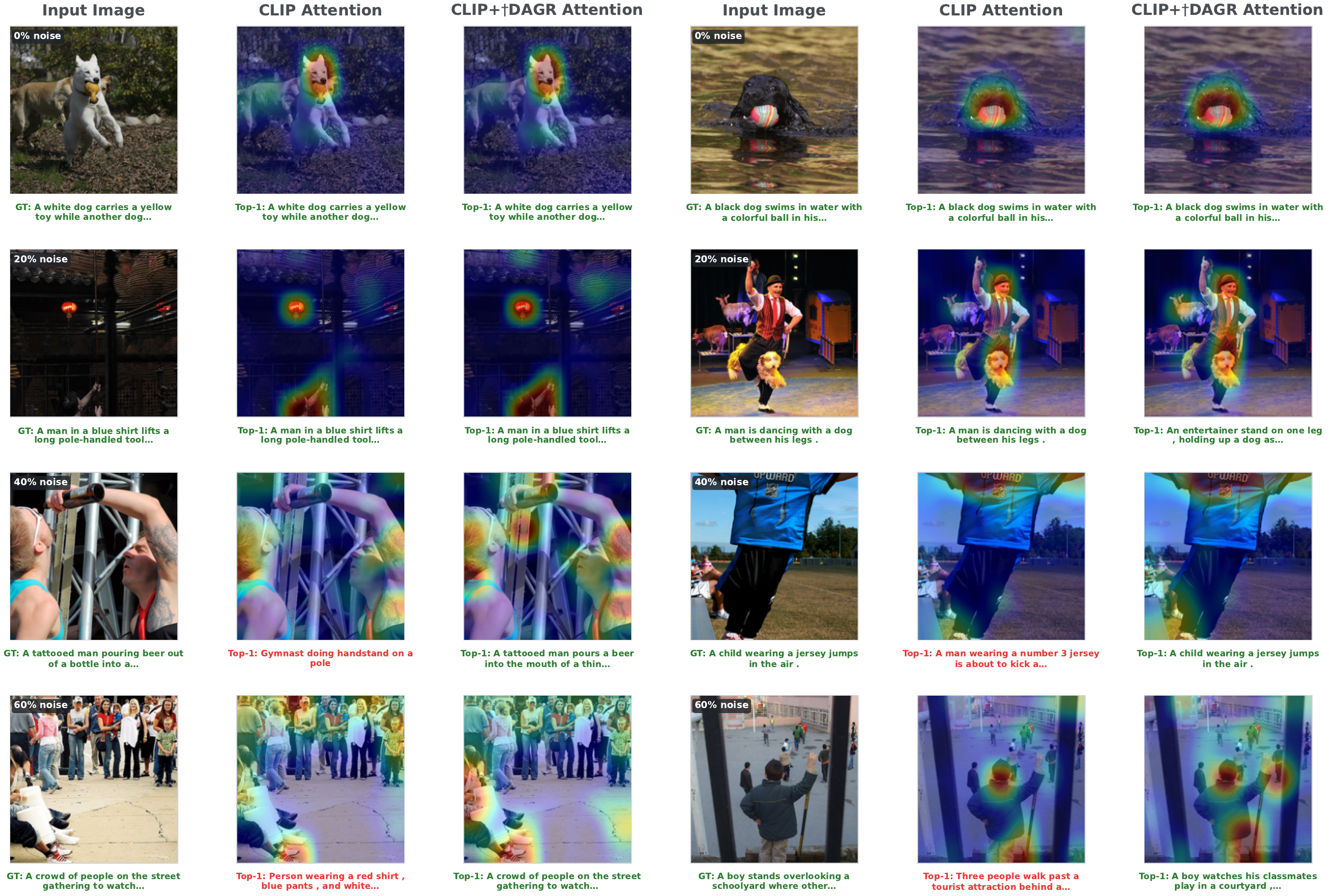}
    \caption{
        \textbf{Qualitative comparison under noisy image--text correspondence.}Each row shows two clean test images evaluated using models trained with the indicated fraction of randomly mismatched image--text pairs. For each image, we report the ground-truth caption and compare the attention maps and top-1 retrieved captions produced by CLIP and CLIP+\regName. Green and red text denote semantically correct and incorrect retrievals, respectively. All attention maps are generated using the same visualization procedure and normalization. These selected examples are intended as qualitative complements to the aggregate retrieval results.
    }
    \label{fig:clip_attention}
\end{figure*}

\paragraph{Qualitative analysis under noisy image--text correspondence.}
Figure~\ref{fig:clip_attention} presents selected clean test images together with attention maps and top-1 retrieved captions from CLIP and CLIP+\regName, where the models are trained with increasing fractions of randomly mismatched image--text pairs. Under clean or moderate corruption, the two models exhibit broadly similar attention patterns and generally retrieve captions consistent with the ground truth, suggesting that \regName does not disrupt the clean visual representation. Under stronger correspondence noise, vanilla CLIP produces semantically mismatched top-1 results in several selected examples, whereas CLIP+\regName retains predictions that better match the depicted objects or interactions. The attention differences are localized and are not uniformly sharper across all examples; we therefore interpret these visualizations as qualitative illustrations rather than independent evidence of improved visual grounding. Together with the aggregate retrieval results, they suggest that \regName yields representations that are less sensitive to corrupted cross-modal supervision.

\end{document}